\documentclass{article}

\usepackage[preprint]{corl_2021} 
\usepackage{times}

\usepackage{adjustbox}
\usepackage{enumitem}
\usepackage{sidecap}

\sidecaptionvpos{figure}{c} 
  
\usepackage{multicol}
\usepackage{url}
\usepackage{booktabs}       
\usepackage{amsfonts}       
\usepackage[utf8]{inputenc}
\usepackage{graphicx}
\usepackage{wrapfig}
\usepackage{lipsum}

\usepackage{caption}
\usepackage{subcaption}     
\usepackage{makecell}       
\usepackage{multirow}       
\usepackage{mdwtab,booktabs}
\usepackage{float}          
\usepackage{amsmath}
\usepackage{amssymb}
\usepackage{bm}             

\newsavebox{\twosubbox}
\newcommand{\R}{\mathbb{R}}

\newcommand{\method}{{ToC}}

\title{Touch-based Curiosity for Sparse-Reward Tasks}

\author{Sai Rajeswar$^{1,2,3}$\thanks{Equal contribution.  Correspondence to: Sai Rajeswar$<$sai.rajeswar.mudumba@umontreal.ca$>$}, Cyril Ibrahim$^{3}$\footnotemark[1], Nitin Surya$^{3}$, \\ \textbf{Florian Golemo$^{1,2,3}$, David Vazquez$^{3}$, Aaron Courville$^{1,2,4}$, Pedro O. Pinheiro$^{3}$} \\
$^{3}$Montréal Institute for Learning Algorithms, $^{2}$Université de Montréal,  \\$^{3}$Element AI, Montréal, $^{4}$CIFAR Fellow 
}

\begin{document}
\maketitle

\begin{abstract}
    Robots in many real-world settings have access to force/torque sensors in their gripper and tactile sensing is often necessary in tasks that involve contact-rich motion. 
In this work, we leverage surprise from mismatches in touch feedback to guide exploration in hard sparse-reward reinforcement learning tasks.
Our approach, Touch-based Curiosity (\method{}), learns what visible objects interactions are supposed to ``feel" like. We encourage exploration by rewarding interactions where the expectation and the experience do not match.
We test our approach on a range of touch-intensive robot arm tasks (e.g. pushing objects, opening doors), which we also release as part of this work.
Across multiple experiments in a simulated setting, we demonstrate that our method is able to learn these difficult tasks through sparse reward and curiosity alone.
We compare our cross-modal approach to single-modality (touch- or vision-only) approaches as well as other curiosity-based methods and find that our method performs better and is more sample-efficient.
\end{abstract}
\section{Introduction}\label{introduction}
\vspace{-2mm}
Most successes in reinforcement learning (RL) come from games~\cite{MnihKSGAWR13,Silver} or scenarios where the reward is strongly shaped~\cite{zhu18,sebastien18}. 
In the former, the environment is often fully observable, and the reward is dense and well-defined. 
In the latter, a large amount of work is required to design useful reward functions. 
While it may be possible to hand-craft dense reward signals for many real-world tasks, we believe that it is a worthwhile endeavor to investigate learning methods that do not require dense rewards.

Closely related to the sparse rewards problem is the issue of exploration. One reason that traditional RL agents struggle with sparse-reward problems is a lack of exploration. An agent may not obtain useful rewards without an intuitive exploration strategy when rewards are sparse. Exploration based on intrinsic curiosity comes naturally to many animals and infants (who start crawling and exploring the environment at around 9 months~\cite{Vernon} and oftentimes even before they can crawl by using their hands and mouth to touch and probe objects). Touch is a local experience and encodes accurate geometrical information while handling objects. Experimental studies in infants has suggested that tactile and visual sensory modalities play a central role in systematic learning of tasks related to object understanding, interaction and manipulation~\cite{touch, psycho71}. 

Ideally, we would like our RL agents to explore the environment in an analogous self-guided fashion to learn the dynamics and object properties, and use this knowledge to solve downstream tasks.   Just as how humans utilize different sensory modalities to explore and understand the world around them, exploration in robots should be more embodied and related to a combination of vision and touch and potentially other sensor modalities. We believe that building autonomous agents that are self-driven and seek to explore via multi-modal interaction are crucial to address key issues in developmental robotics.


Recent works in RL have focused on a curiosity-driven exploration through prediction-based surprise~\cite{Burda19, Pathak, Raileanu}. 
In this formulation, a forward dynamics model predicts the future, and if its prediction is incorrect when compared to the real future, the agent is surprised and is thus rewarded. 
This encourages the agent to look for novel states while improving its visual forward model in return.
However, this formulation can be practically challenging to optimize since there are many states that are visually dissimilar but practically irrelevant (e.g. for a pushing task, moving a robotic end-effector without touching the object creates visual novelty but contributes little to task-related knowledge). 
One way to constrain this search space over curious behaviors is by involving another modality like touch. 

In this work, we demonstrate that a self-guided cross-modal exploration policy can help solve sparse-reward downstream tasks that existing methods without this curiosity struggle to solve.
Our method uses cross-modal consistency (mismatch between visual and touch signal) to guide this exploration.
To use self-play knowledge in downstream tasks, we relabel past experiences, providing a dense reward signal that allows modern off-policy RL methods to solve the tasks.
While there are many existing methods that use artificial curiosity/intrinsic motivation, the majority of these methods either rely on strong domain knowledge (e.g. labels of state dimensions in \citet{sebastien18}, a goal-picking strategy in \citet{Her}) or are prone to get stuck in local optima when a single meaningless stimulus creates enough surprise to capture the attention of the agent (e.g. noisy-TV experiment from~\cite{Burda}). 
Other approaches depend on unrealistic assumptions and goal conditioning~\cite{Her}. 
Our method presents a novel approach in the family of prediction-based models~\cite{Burda19, sandy19, Pathak} and yields better performance on a wide range of robotic manipulation tasks than purely vision-based and touch-based approaches~\cite{Burda19, Pathak}. The tasks are chosen with careful consideration---they comprise of preliminary robotic manipulations such as grasping, pushing, and pulling. 
In this work, we present the following contributions:
\begin{itemize}
    \item A new curiosity method to help solve sparse-reward tasks that use cross-modal consistency (predicting one modality from another) to guide exploration. 
    We implement it in this work as vision and touch modalities, but the formulation of our method does not require any knowledge about the underlying modalities and can thus be applied to other settings.
    \item We create and maintain a manipulation benchmark of simulated tasks, \emph{MiniTouch}, inspired by \citet{chen2020batch,Her}, where the robotic arm is equipped with a force/torque sensor. This allows evaluation of models' performance on different manipulation tasks that can leverage cross-modal learning. 
    \item We validate the performance of our method on MiniTouch environment comprising of four downstream tasks. We compare purely vision-based curiosity approaches and standard off-policy RL algorithms. Our method improves both performance and sample efficiency. 
\end{itemize}

\section{Related Work}
\vspace{-2mm}
\paragraph{Intrinsic Motivation}
Intrinsic motivation is an inherent spontaneous tendency to be curious or to seek something novel in order to further enhance one's skill and knowledge~\cite{white59, Barto04,intrinsic17}. This principle is shown to work well even in the absence of a well-defined goal or objective.
In reinforcement learning, intrinsic motivation has been a well-researched topic~\cite{Schmidhuber, Oudeyer, Oudeyer9, Lair, Marino, Savinov}.
An intuitive way to perform intrinsic motivation is through the use of ``novelty discovery''. For example, incentivize the RL agent to visit unusual states or states with substantial information gain~\cite{HouthooftCDSTA16}. In its simplest form, this can be achieved with up-weighting mechanisms such as state visitation counts~\cite{Strehl}. Count-based methods have also been extended to high-dimensional state spaces~\cite{Bellemare,Burda,Ostrovski}. 
Alternative forms of intrinsic motivation include disagreement~\cite{sekar2020planning}, empowerment~\cite{Klyubin, GregorRW16}. 

Exploratory intrinsic motivation can also be achieved through ``curiosity''~\cite{Schmidhuber, DUBEY2020}.
 In this setting, an agent is encouraged to visit states with high predictive errors~\cite{Raileanu, Burda19, Pathak} by training a forward dynamics model that predicts the future state given the current state and action. Instead of
making predictions in the raw visual space,~\citet{Pathak} mapped images to a feature space where relevant information is represented via an inverse dynamics model.~\citet{Burda19} demonstrate that random features are sufficient for many popular RL game benchmarks. This approach may work well with tasks that require navigation to find a reward because each unseen position of the agent in the world leads to high intrinsic reward when unseen. However, in the case of manipulative tasks, we are less interested in the robot visiting all the possible states and more interested in states where the robot interacts with other objects.
In this work, we leverage multimodal inputs that encourage the agent to find novel combinations of visual and force/torque modalities. 

\paragraph{Self-Supervised Learning via Cross-modality}
Exploiting multimodality to learn unsupervised representations dates back to at least 1993~\cite{deSa94multimodal}. Multimodal signals are naturally suitable for self-supervised learning, as information from one modality can be used to supervise learning for another modality. 
Different modalities typically carry different information, e.g., visual and touch sensory modalities emerge concurrently and often in an interrelated manner during contact-rich manipulation tasks~\cite{synergy}.
Specifically, force/torque motor signal has always been a major component in the literature of perception and control~\cite{Kalakrishnan, levine16,pmlr-liu17a}.

The most common ways to leverage multimodal signals to learn representations are through vision and language~\cite{Srivastava, zhe20}, or through visual and audio~\cite{Gao, owens2018audiovisual}. \citet{Gao_icra, Li_2019_CVPR} demonstrated a unified approach to learning representations for prediction tasks using visual and touch data.
In robotics and interactive settings, the use of additional modalities such as tactile sensing~\cite{Hoof16, Calandra, Murali}  is increasingly popular for grasping and manipulation tasks.~\citet{Lee} showed the effectiveness of self-supervised training of tactile and visual representations by demonstrating its use on a peg insertion task. 
 
While the mentioned approaches have used multiple sensory modalities for learning better representations, in this work we demonstrate its utility for allowing agents to explore.
Similar to ours,~\citet{dean2021see} use multimodal sensory association (i.e. audio and visual) to compute the intrinsic reward. Their curiosity-based formulation allows an agent to efficiently explore the environment in settings where audio and visual signals are governed by the same physical processes. In addition to the different nature of sensory signals, they use a discriminator that determines whether an observed multimodal pair is novel. This might not work in our case as touch is a more sparse signal and using a discriminator could lead to ambiguous outcomes. 
\vspace{-1mm}
\begin{figure*}
    \centering
     \begin{subfigure}{0.66\textwidth}
    \includegraphics[width=\linewidth]{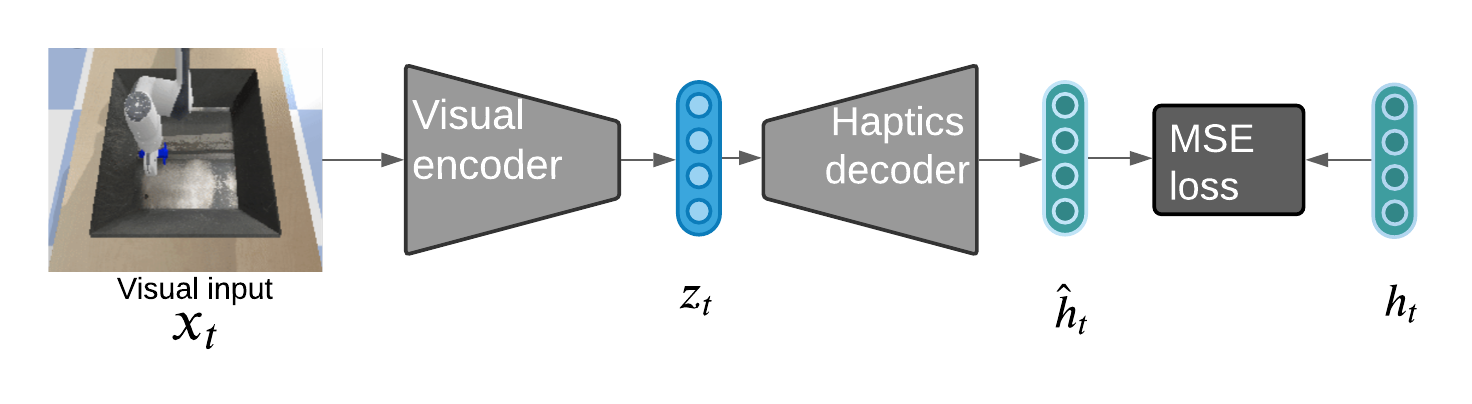}
    \caption{Touch Control module} \label{fig:1a}
    \end{subfigure}
    \begin{subfigure}{0.3\textwidth}
    \includegraphics[width=\linewidth]{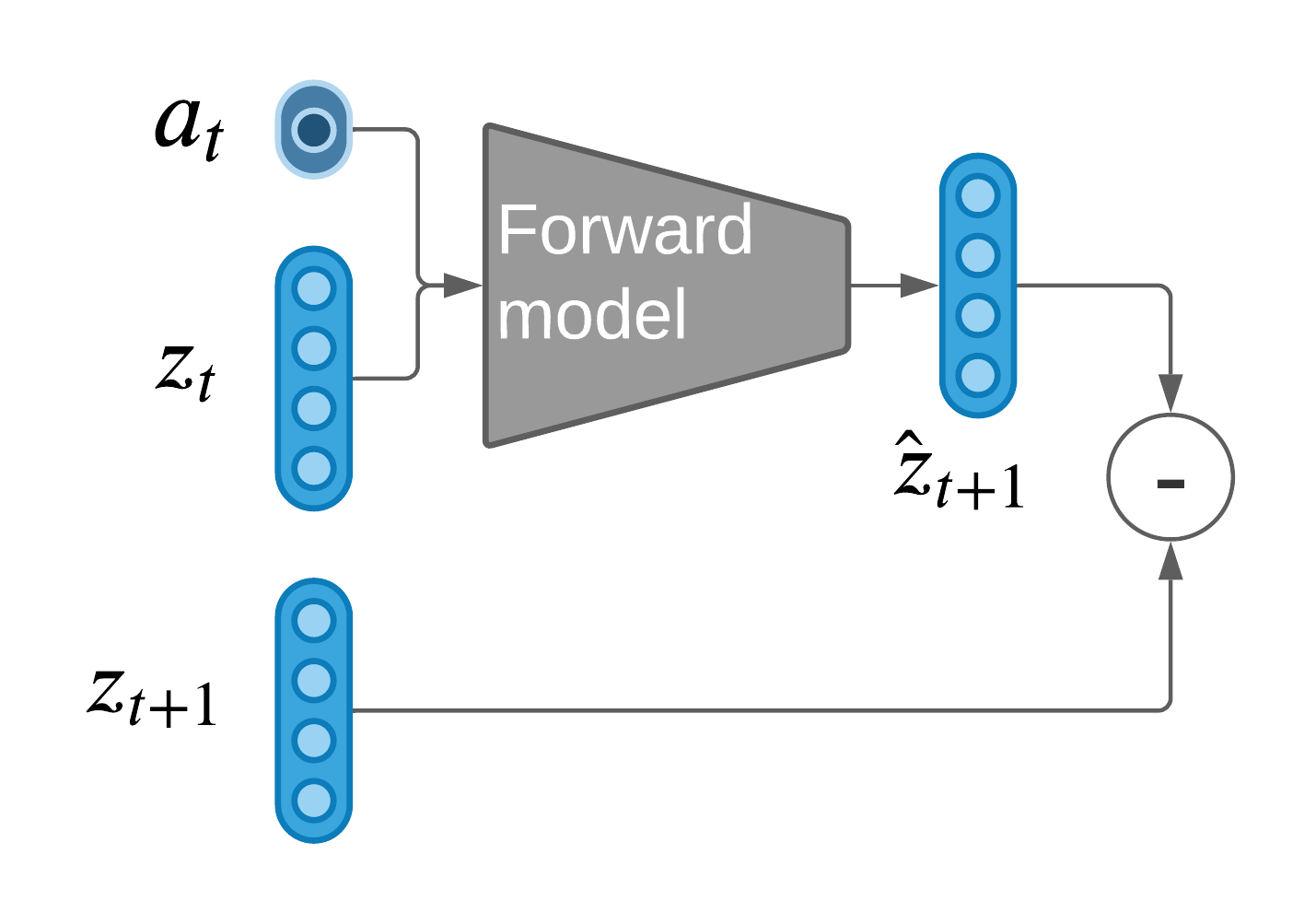}
    \caption{Forward module} \label{fig:1b}
    \end{subfigure}%
    \caption{\textbf{Touch-based Curiosity Model.} (a) The input image $x_t$ at time $t$ is transformed into a 256-dimensional feature vector $z_t = enc(x_t)$ using a  CNN encoder. The touch decoder network predicts corresponding force/torque vector $\hat{h}_t = dec(z_t)$. The $L2$ norm between predicted haptics $\hat{h}_t$ and observed haptics $h_t$ is used as exploration reward. (b) To stabilize training, an additional network is used to predict the forward dynamics, and the difference between predicted next latent state $\hat{z}_{t+1}$ and actual next latent state $\hat{z}_t$ is used as weighted additional term in the exploration reward.}
    \label{fig:model-figure}
\end{figure*}
\vspace{-1mm}
\section{Proposed Approach}

\vspace{-1mm}
The goal of our method, Touch-Based Curiosity (\method{}),  is to encourage the agent to interact with objects. \method{} provides a reward signal for a RL agent to explore the state space of a task that involves interacting with objects.
The exploration phase is independent of the downstream task, i.e., relying solely on visual and force/torque signals, without a reward signal from the downstream task.

Similar to how people spend more time exploring stimuli that are more incongruous~\cite{psycho71}, \method{} guides the agent to focus its experience on different novel cross-modal associations.
We augment this intrinsic objective with a future visual state prediction objective similar to the one in \citet{Pathak} to avoid getting stuck in undesired inactive configurations.
Note that we sometimes refer to the future state prediction objective in the text as forward dynamics objective.
In this work, we focus on the cross-modality between vision and touch, but the same idea could be applied to other pairs of sensory domains, such as vision and sound, or touch and acoustics.
\vspace{-2mm}
\subsection{Problem Formulation}
\vspace{-1mm}
The learning problem is formalized as a Markov decision process (MDP) defined by a
tuple $\{\mathcal{S}, \mathcal{A}, \mathcal{T} , \mathcal{R}, \rho, \gamma\}$ of states, actions, transition probability, reward, initial state distribution, and
discount factor. The goal is to find the optimal policy that maximizes the discounted sum of rewards, $\pi^{*} = \mathbb{E}_\pi[\sum_{t}^{\infty} \gamma^t r(\mathbf{s}_t, \mathbf{a}_t)]$. In our case, each state $s_t$ in the trajectory is composed of both visual and corresponding touch features as detailed in the  following section. 
 We use soft actor-critic policy gradients (SAC)~\cite{haarnoja2018soft} to train our policies, but in principle, our proposed approach is algorithm-agnostic.
The policy $\pi$ is evaluated with an estimation of the soft Q-value:
\begin{equation}
\label{eqn:q_value}
Q(\mathbf{s}_t, \mathbf{a}_t) \triangleq r(\mathbf{s}_t, \mathbf{a}_t) + \gamma\mathbb{E}_{\mathbf{s}_{t+1}\sim p}[V(\mathbf{s}_{t+1})] \;, 
\end{equation}
where $V(\mathbf{s}_t) = \mathbb{E}_{\mathbf{a}_t\sim \pi} [ Q(\mathbf{s}_t, \mathbf{a}_t) - \log \pi (\mathbf{a }_t | \mathbf{s}_t) ]$ is the soft value function.

\subsection{Touch-based Prediction}

Our core prediction framework consists of two modules: (i) a  \textit{touch control} module, which learns to predict expected touch signal from the visual input, and (ii) a \textit{forward dynamics} model, which predicts the next latent state from the current latent state and the current action (see Figure \ref{fig:model-figure}). 
Let the state of the environment $\mathbf{s}_t=(z_t, h_t)$ at time $t$ be composed of a visual feature $z_t$ (the encoded visual input) and a touch signal $h_t$.
The touch prediction model consists of a convolutional encoder $z_t = enc(x_t)$ that transforms the image $x_t$ into a latent representation $z_t$ and a fully-connected decoder $\hat{h}_t = dec(z_t)$ that transforms the latent into a predicted haptic signal $\hat{h}_t$.
The encoder-decoder is trained with a L2 reconstruction loss, i.e. for every image $x_t$ and force/torque sensor $h_t$:
\begin{equation} \label{eq:loss-touch}
L_{touch} = \left \| \hat h_t - h_t \right \|_2 \;.
\end{equation} 

A high prediction error on a given image indicates that the agent has had few interactions like this. 
Therefore, to harness this ``surprise" to guide exploration, we define the intrinsic reward at time $t$ during exploration to be proportional to this reconstruction loss. This essentially allows the agent policy to visit under-explored configurations of the state space by encouraging interactions where the system does not know what the target object ``feels'' like. In addition to  efficient exploration, being aware of such incongruity via touch prediction aids learning local regularities in the visual input. This in turn could assist  better generalization to unseen states.  An overall pipeline of the framework is shown in Figure~\ref{fig:1a}.
\vspace{-2mm}
\subsection{Regularization Through Forward Dynamics Model}
 \vspace{-1mm}
We found empirically (and we demonstrate in the experiments section below) that the surprise stemming from haptic novelty was not enough to cause object-centric interaction. 
We postulate that by incorporating  visual surprise (i.e. the mismatch between predicted forward dynamics and observed dynamics)~\cite{Pathak}, we can create an agent that seeks out visual novelty as well as haptic one and thus leads to better state space coverage.
To this end, our model is augmented with a forward dynamics model (see Figure-\ref{fig:1b}) $\hat{z}_{t+1} = fdm(z_t,a_t)$ that learns to map the latent state $z_t$ (obtained from the visual encoder $enc$) and action $a_t$ at time $t$ to the predicted latent state $\hat{z}_{t+1}$ at the next timestep.
This model is also trained with L2 loss:
\begin{equation} \label{eq:loss-fut}
\begin{aligned}
L_{fdm} &= \left \| \hat{z}_{t+1} - z_{t+1} \right \|_2 \; =  \left \| fdm(enc(x_t),a_t) - enc(x_{t+1}) \right \|_2 \;.
\end{aligned}
\end{equation}

The intrinsic reward is defined as the convex combination of the cross-modal prediction loss (Eq.~\ref{eq:loss-touch}) and the forward dynamics model loss (Eq.~\ref{eq:loss-fut}):
\begin{equation} \label{eq:final-reward}
\begin{aligned}
 r_t = (1-\lambda) \cdot L_{touch} + \lambda \cdot L_{fdm}\;,
\end{aligned}
\end{equation}
where $\lambda \in[0,1]$ is a balancing factor. The effect of the factor $\lambda$ on overall performance is studied in the ablation experiments described in Section~\ref{forward_ablation}. 
\vspace{-2mm}
\subsection{Training}
\vspace{-1mm}
Learning is divided into two stages: (i) an \emph{exploratory} step, where the agent performs free exploration following \method, and (ii) an \emph{adaptation} step, where the agent is tasked to solve a downstream problem, given a sparse reward.

During the exploratory step, each trajectory consists of pairs of image and force/torque features, $(z_1$, $h_1)$, $(z_2$, $h_2)$,..., $(z_n$, $h_n)$. These trajectories are used for two purposes: (i) updating the parameters of the prediction model to help shape the representations  and (ii) updating the exploration policy based on the intrinsic reward $r_t$. Note that we use the touch features only to craft the intrinsic rewards and the input to RL agent consists of  visual features alone. For vision-based curiosity models, \citet{Burda19} observed that encoding visual features via a random network constitute a simple and effective strategy compared to learned features. In Section \ref{ablation}, we investigate the performance of our model in both  scenarios, i.e., when the features are learned vs random. The overall optimization problem at this step consists of the policy learning (driven by intrinsic reward), the touch reconstruction loss (Eq.~\ref{eq:loss-touch}), and the forward dynamics loss (Eq.~\ref{eq:loss-fut}).
During the downstream adaptation step, the parameters of the policy network, the Q network and the replay buffer are retained from the exploratory phase. The objective of the down-stream task is computed as:
\begin{equation} 
\underset{\theta}{\min}
\begin{bmatrix}
  -\mathbb{E}_{\pi}\begin{bmatrix}\sum _t r_t^e \end{bmatrix} 
  \end{bmatrix}\;,
\end{equation}
where $r_t^e$ in this phase is task-specific external sparse reward. In both steps, the objectives are optimized with Adam~\cite{KingmaB14}. 
\begin{figure*}
    \centering
    \includegraphics[width=0.42\textwidth]{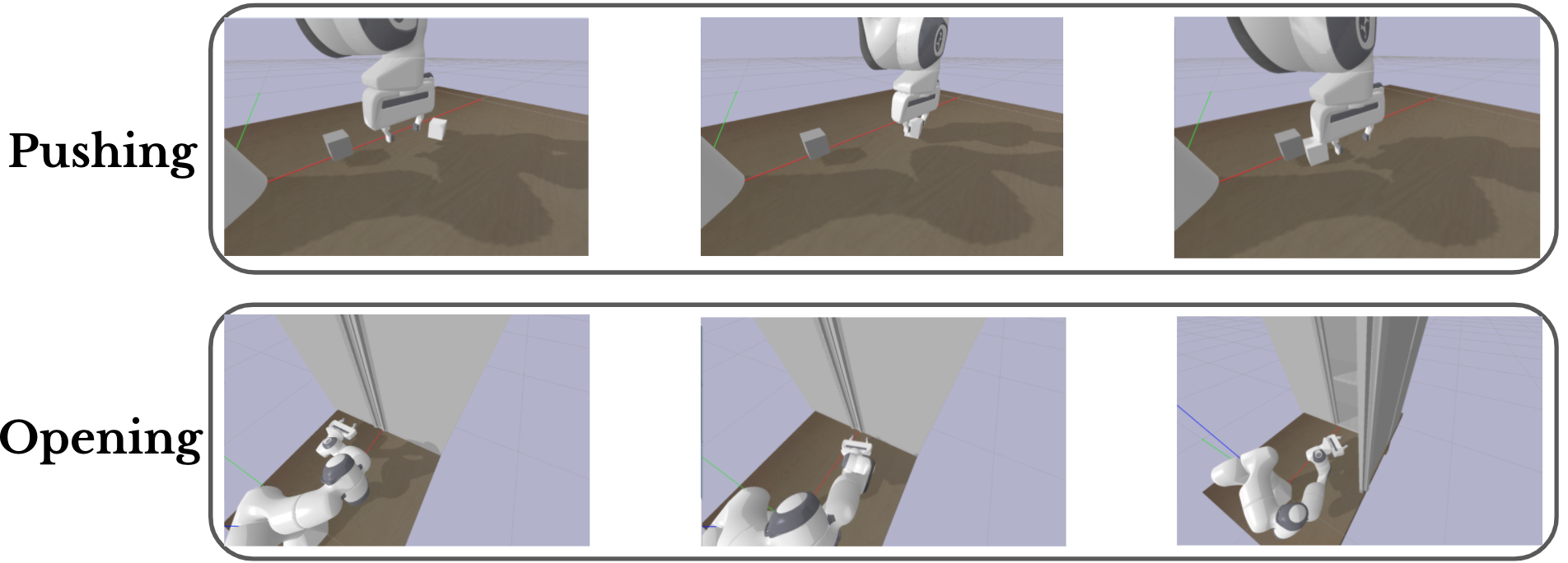}
    \includegraphics[width=0.42\textwidth]{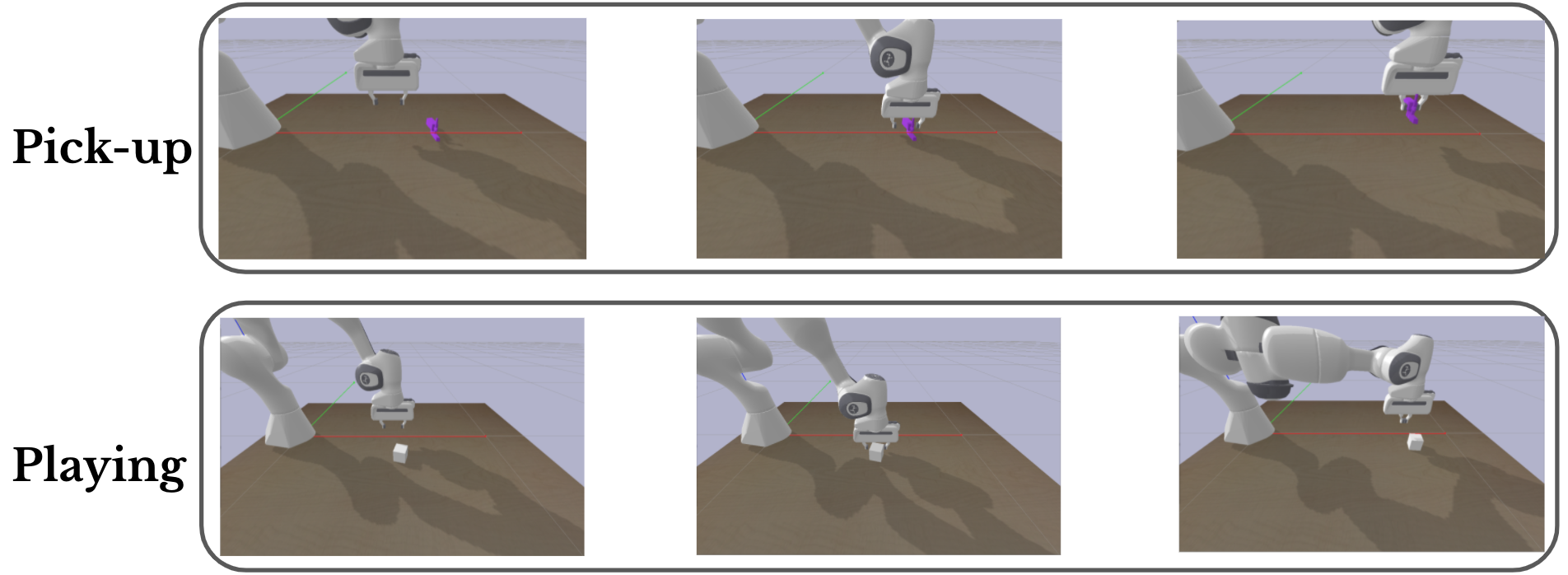}
    \caption{\textbf{`MiniTouch' benchmark tasks}  include opening a door, pushing an object to a target, grasping and lifting an object and a toy task in which object interactions are counted. The figure shows consecutive frames from a successful episode in each task.}
    \label{fig:test-tasks}
\end{figure*}

\textbf{Implementation Details}: 
The encoder is a four-layered strided CNN followed by a fully connected network. We use LeakyReLU~\cite{XuWCL15} as non-linear activation in all the layers. The decoder network is a two-layerd MLP that maps 256-dimensional visual features to touch vectors.  
For a more detailed description of the networks (SAC policy network and \method{} networks, including the forward model) and hyper-parameters, please refer to the supplementary material Section \ref{suppl_experimental}.


\section{Experimental Setting}

Our experiments focus on a tabletop robot manipulation from raw image observations and raw force/torque sensory values, which we refer to as ``touch vector''. 
For our experiments, we use a 7-DoF Franka Emika Panda arm with a  two-finger parallel gripper. 
Each of the fingers is equipped with a simulated force/torque sensor that measures the joint reaction force applied to it.  
We utilize PyBullet~\cite{coumans2019} to simulate the robot arm and haptic sensor\footnote[1]{Based on code from the official Franka Emika repo https://github.com/frankaemika/libfranka}. 

\subsection{MiniTouch Benchmark}

Our proposed benchmark, MiniTouch, consists of four manipulation tasks: Playing, Pushing, Opening, Pick-up. Each of the tasks along with  corresponding actions, observations, and rewards is  described in detail in the Supplementary Material Section \ref{sec:benchmark_sec} and further illustrated in Figure ~\ref{fig:test-tasks}. 
MiniTouch is an active repository\footnote[1]{https://github.com/ElementAI/MiniTouch} and we expect to update the benchmark with new tasks and datasets.
The tasks are inspired by \citet{metaworld} but are \textit{not} based on a proprietary simulator, feature an arm that we have access to for real-world experimentation (in follow-up work), and where the arm is equipped with a haptic sensor.

\subsection{Baseline Comparisons and Metrics}

For the task evaluation, we study two versions of our model:  (i) \method{}-Pure considering the touch vector reconstruction intrinsic reward alone,  and (ii) \method{}, considering the full intrinsic reward (Eq.~\ref{eq:final-reward}).
We compare our model with SAC and several well-known intrinsic exploration based baselines:
\begin{itemize}[noitemsep,nolistsep]
    \item  \textbf{SAC}: The unmodified Soft Actor-Critic algorithm from \citet{haarnoja2018soft}. 
 \item \textbf{ICM}: SAC augmented with the state-of-the-art visual curiosity approach Intrinsic Curiosity Module (ICM)~\cite{Burda}, which uses a visual forward model to guide exploration. 
  \item \textbf{Disagreement}: It  uses model disagreement as objective for exploration~\cite{pathak19a,sekar2020planning}. It leverages variance in the prediction of an ensemble of latent dynamics models as the reward. \item \textbf{RND}: Random Network Distiallation~\cite{rnd} utilizes a randomly initialized neural network
to specify an intrinsic reward for visiting unexplored states in hard exploration problems.
 \end{itemize}

We built a Pytorch~\cite{pytorch} version of these baselines based on their open source code (details in supplementary material). We use the following metrics to evaluate our method and the baseline models:

 \textbf{Exploration success:} measures the percentage of times that the agent attained the goal state in the \emph{exploratory} phase, i.e. with no external reward. Higher is better.\\
 \textbf{Success:} denotes the percentage of times that the agent attained the goal state during the \emph{down-stream} task phase. \\
 \textbf{Episode steps:} quantifies the number of steps required for each episode to succeed. This metric is an indicator of sample efficiency. The lesser the number of steps, the faster the agent's ability to succeed.\\
 \textbf{Touch-interaction:} Amount of interaction the agent's fingers have with the underlying object. We measure this by computing the variance of force/torque sensory signal across the whole episode. Higher variance indicates better interaction. \\
 \textbf{Object movement:}  The agent can resort to constantly engaging with the object in a passive manner in order to satisfy the objective. We, therefore, compute the variance of door angle (for the Opening task) and the variance of object position (for the remaining tasks) over the course of training. A higher motion indicates diverse state space and that the agent does not get stuck with pointless subtle movements of the objects.

\begin{figure}[t]
\begin{subfigure}{0.21\linewidth}
\centering
\includegraphics[width=0.99\columnwidth]{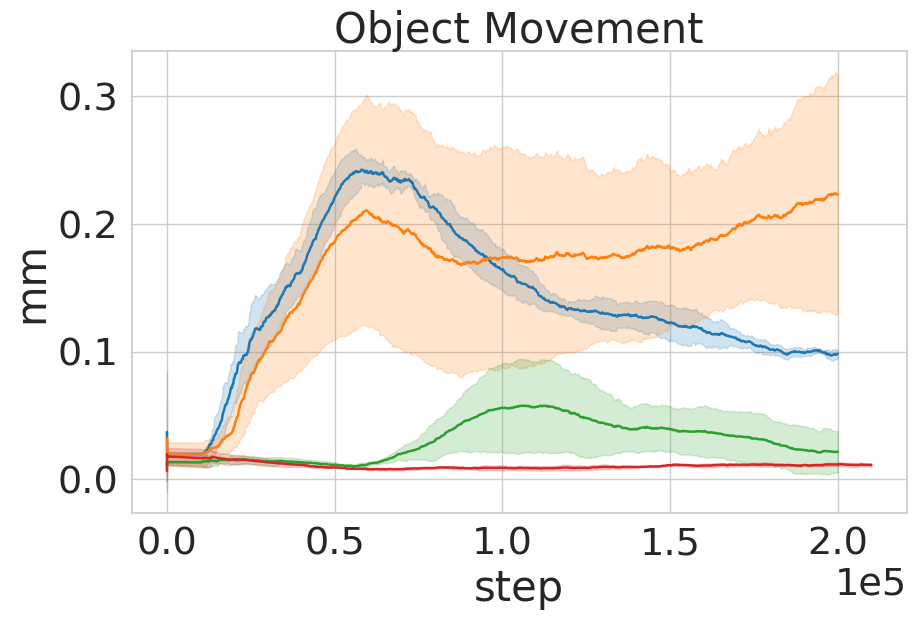}
\caption{}
\label{fig1:sub1}
\end{subfigure}%
\begin{subfigure}{0.22\linewidth}
\centering
\includegraphics[width=0.99\columnwidth]{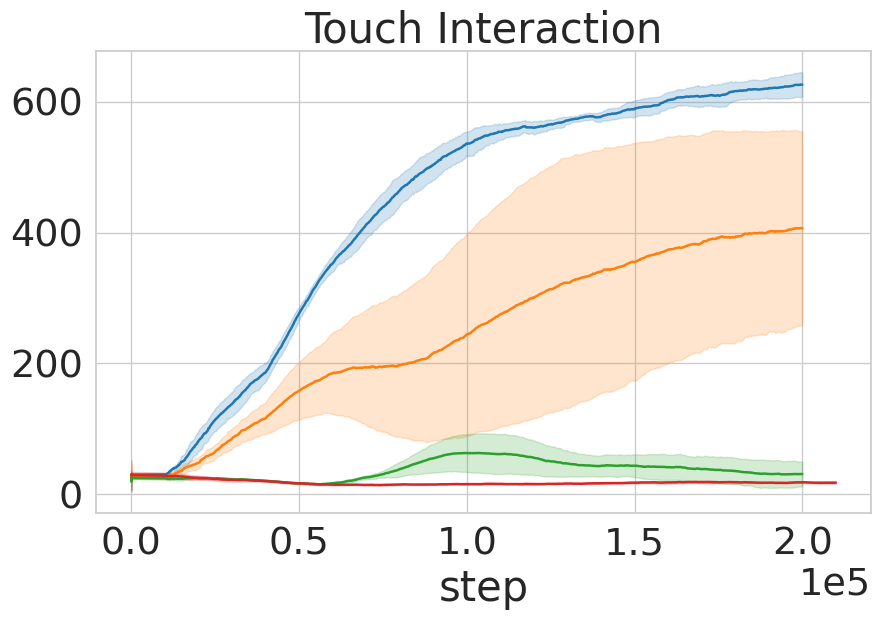}
\caption{}
\label{fig1:sub2}
\end{subfigure}
\begin{subfigure}{0.22\linewidth}
\centering
\includegraphics[width=0.9\columnwidth]{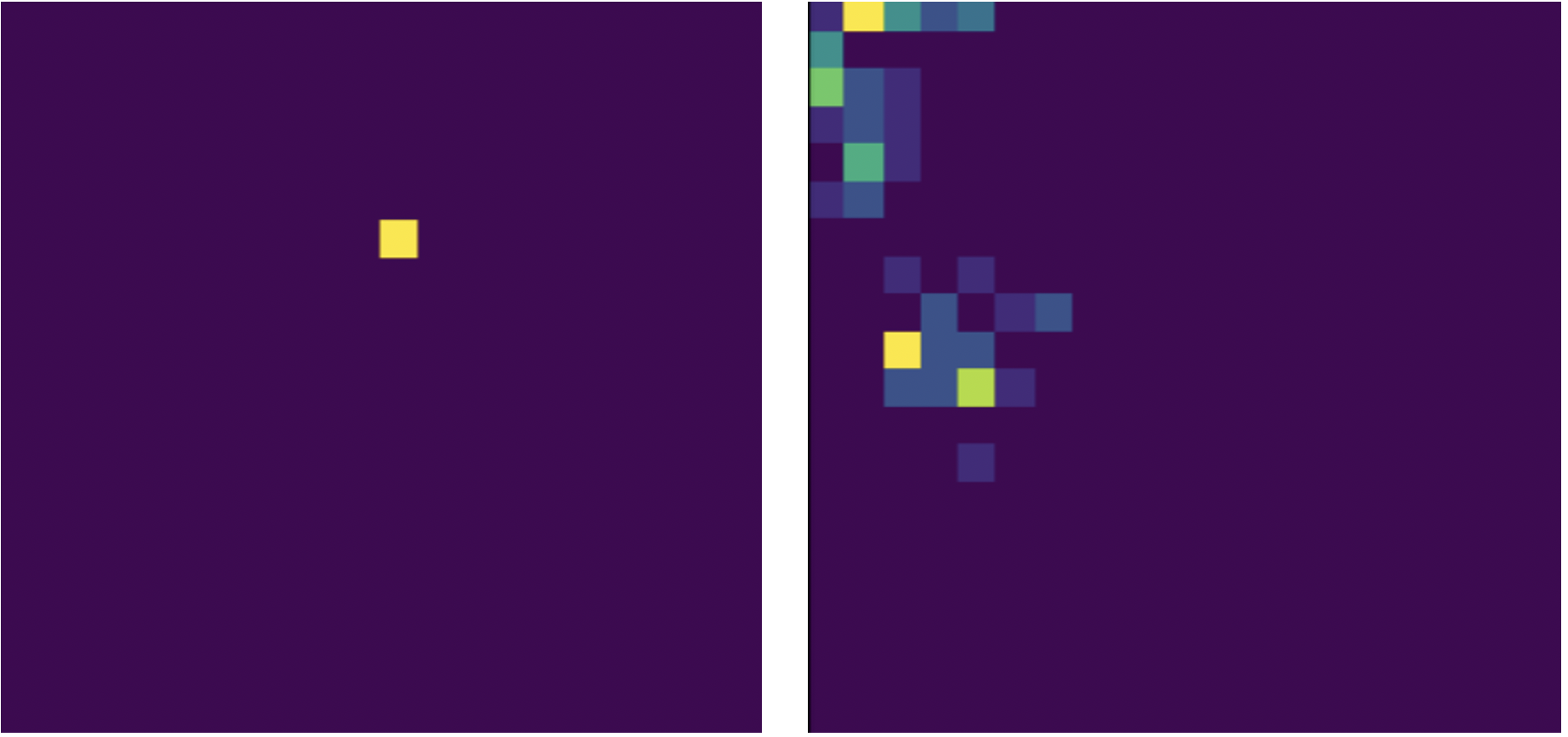}
\caption{}
\label{fig1:sub1_heatmap}
\end{subfigure}%
\begin{subfigure}{0.21\linewidth}
\centering
\includegraphics[width=0.9\columnwidth]{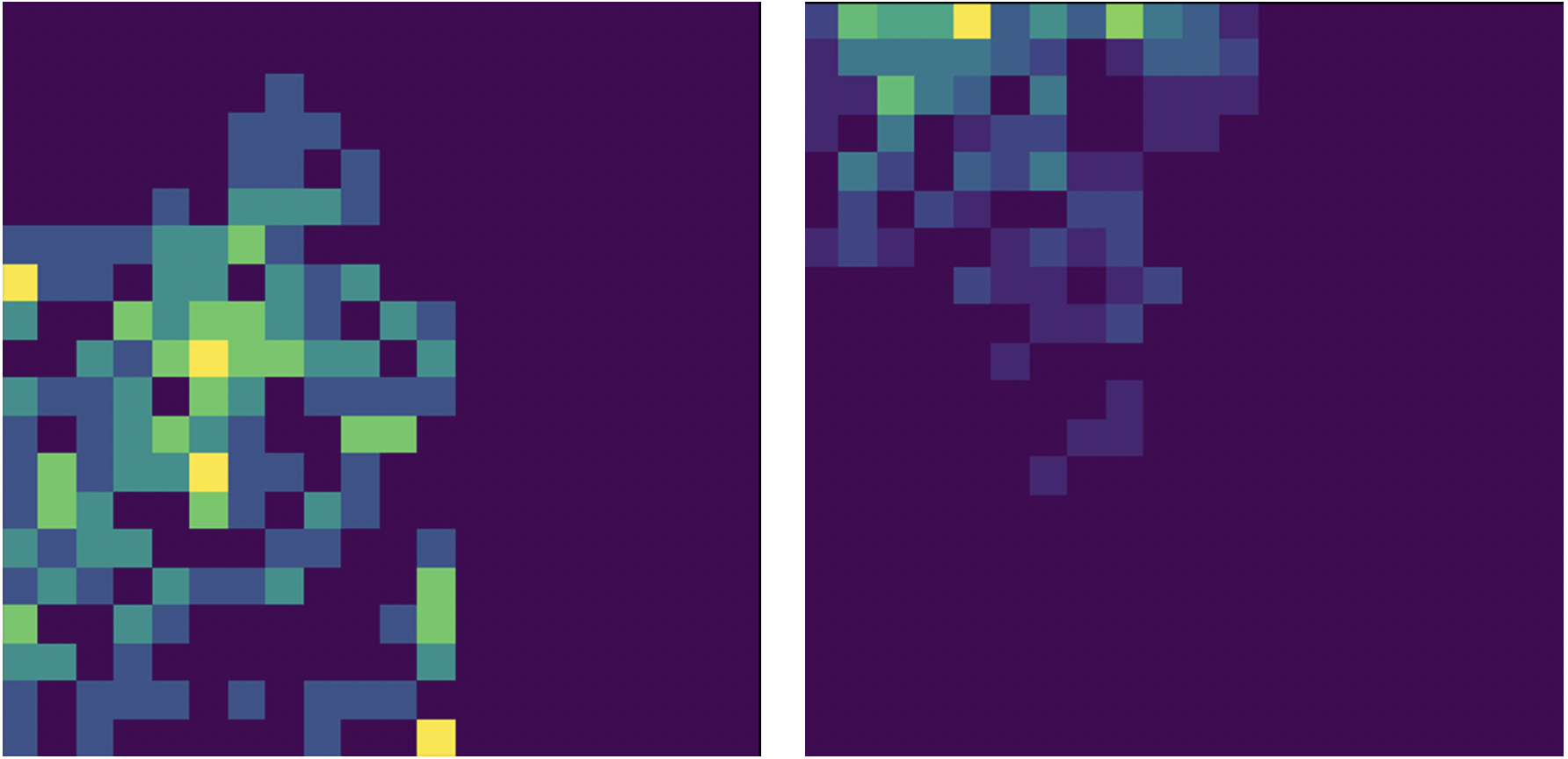}
\caption{}
\label{fig1:sub2_heatmap}
\end{subfigure}\\
\centering
\begin{subfigure}{0.5\linewidth}
\centering
  \includegraphics[width=0.9\columnwidth]{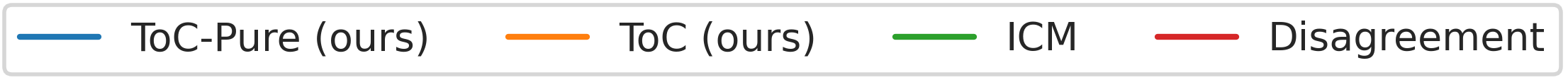}
  \end{subfigure}%
\caption{\textbf{Object Interaction for Playing task.} To quantitatively evaluate an agent's object interaction, we consider both touch interaction and object displacement. (a) shows the object movement evaluation (avg displacement in mm from the starting point) of our method compared with the baseline methods over training steps; (b) depicts the number of touch interactions, evaluated on the ``playing" task. (c) Heat maps showing the object position  at the beginning (\emph{left}) and end (\emph{right}) of the ToC training. The object is moving more often towards the end. (d) shows heat maps showing the end-effector position at the beginning (\emph{left}) and end (\emph{right}) of the training. Robot's fingers are spread every where at starting while learn to focus near the object towards the end of training.}
\label{fig:single_object_variance}
\end{figure}


\begin{figure}[t]
  \centering
  \includegraphics[width=0.07\columnwidth]{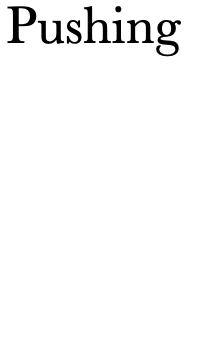}  \includegraphics[width=0.22\columnwidth]{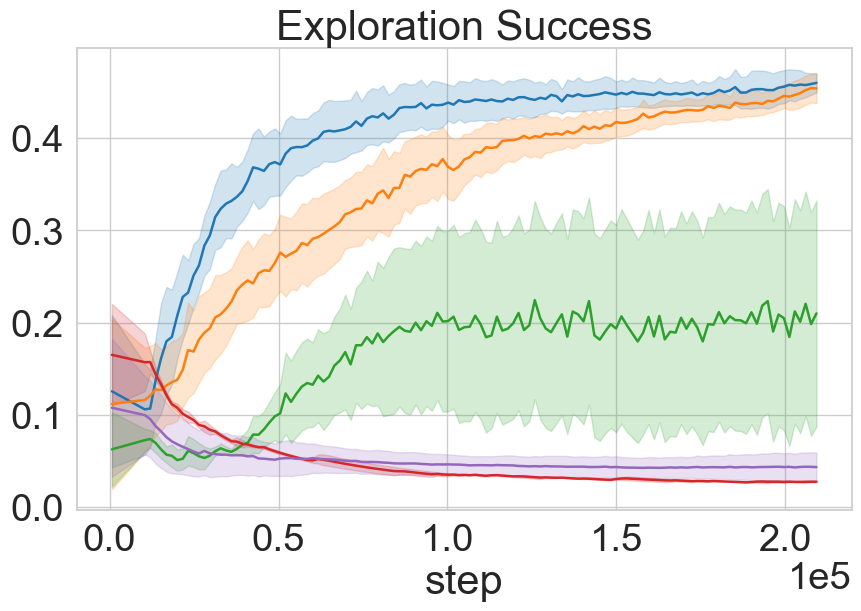}
  \includegraphics[width=0.22\columnwidth]{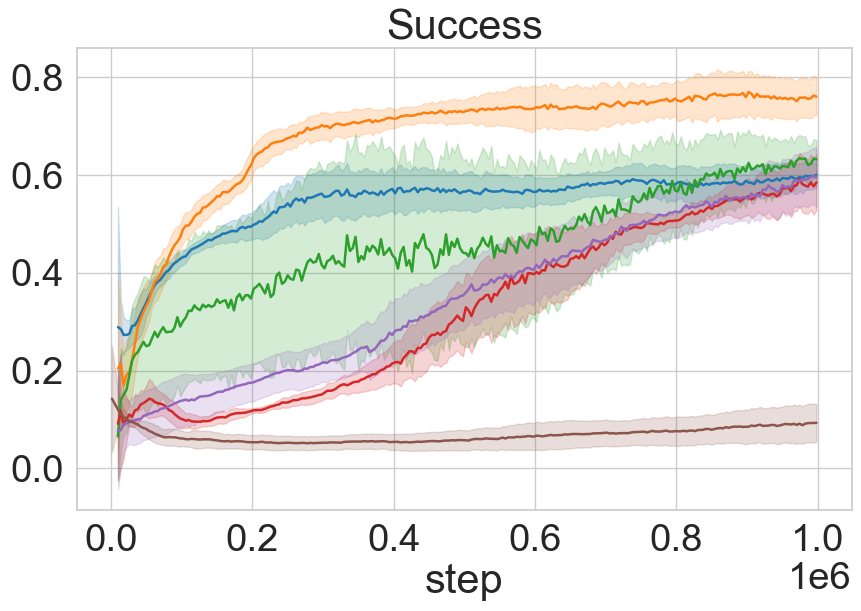}
  \includegraphics[width=0.22\columnwidth]{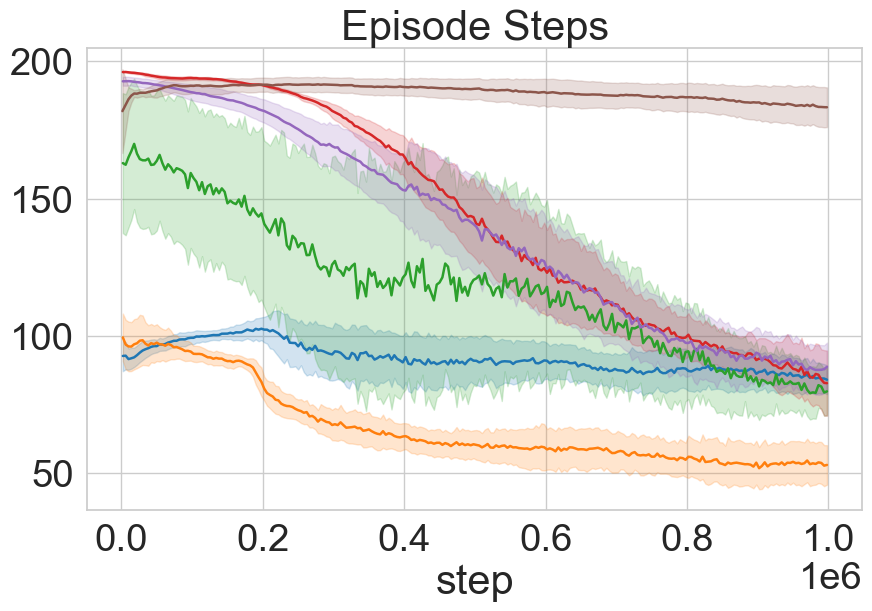}
  \includegraphics[width=0.22\columnwidth]{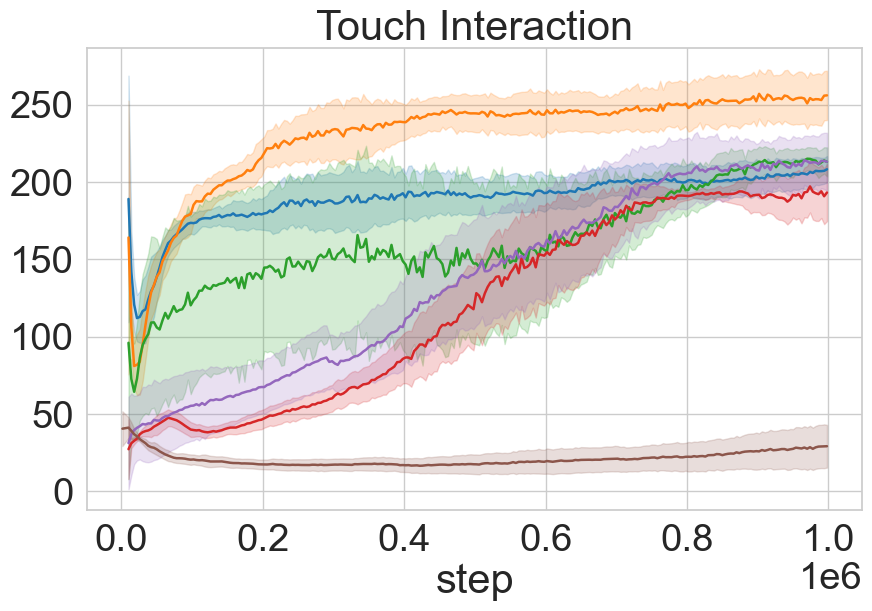}\\
  \includegraphics[width=0.07\columnwidth]{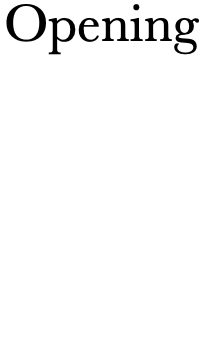}
  \includegraphics[width=0.22\columnwidth]{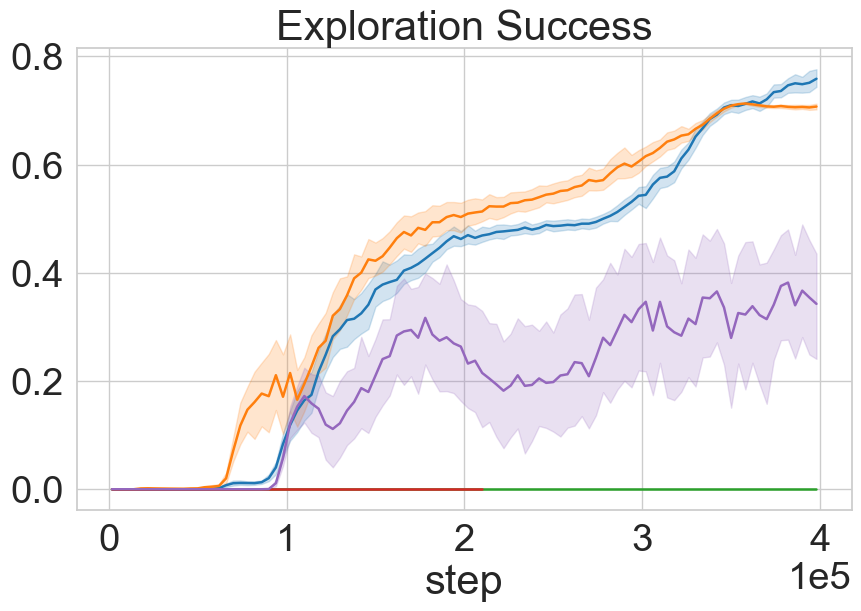}
  \includegraphics[width=0.22\columnwidth]{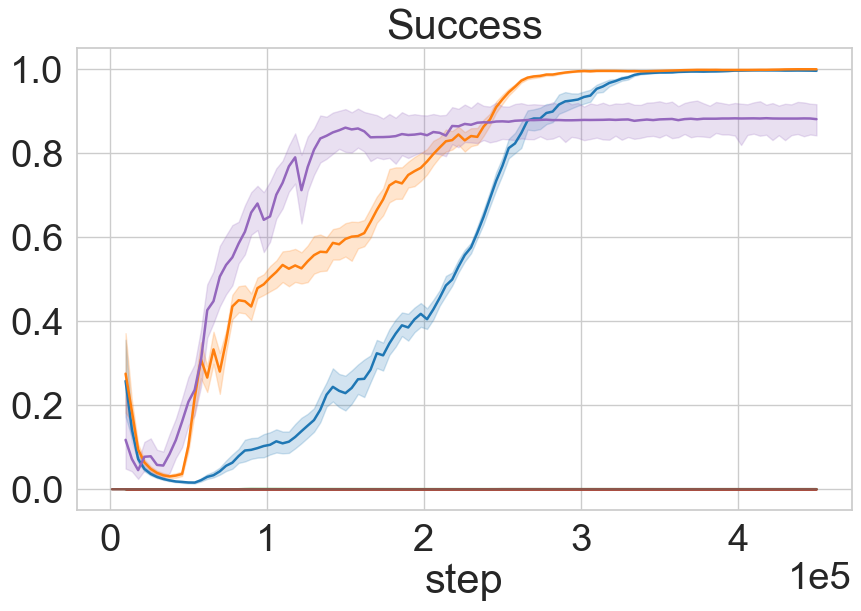}
  \includegraphics[width=0.22\columnwidth]{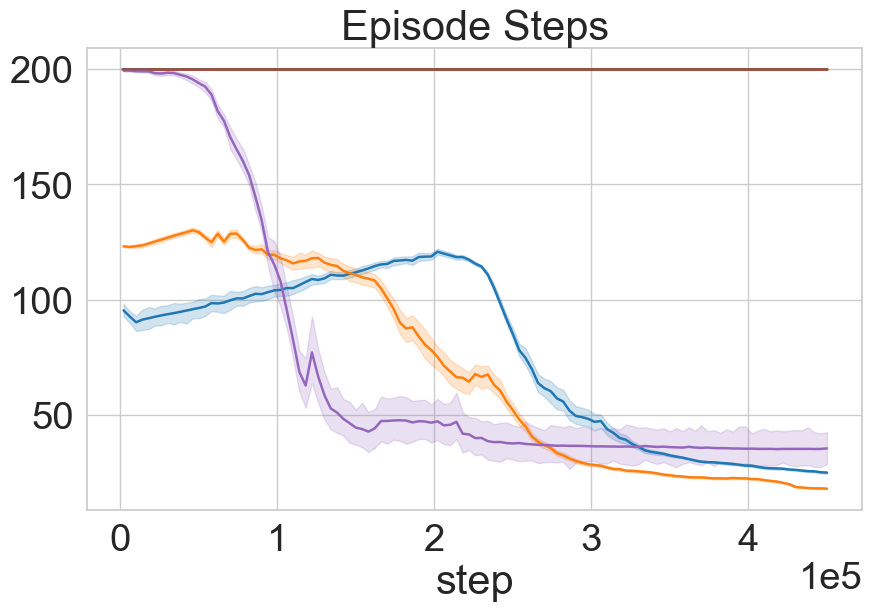}
  \includegraphics[width=0.22\columnwidth]{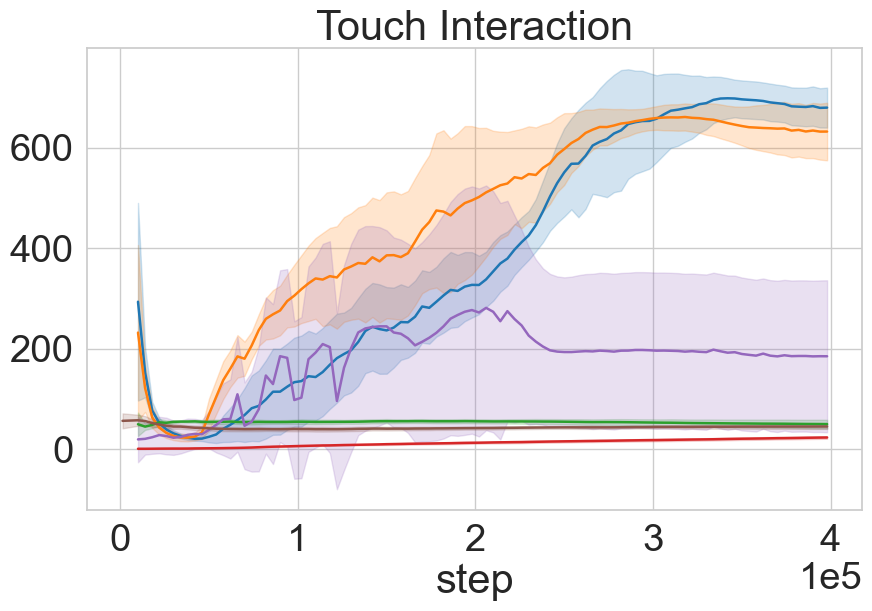}\\
  \includegraphics[width=0.07\columnwidth]{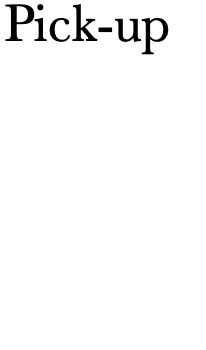}
\includegraphics[width=0.22\columnwidth]{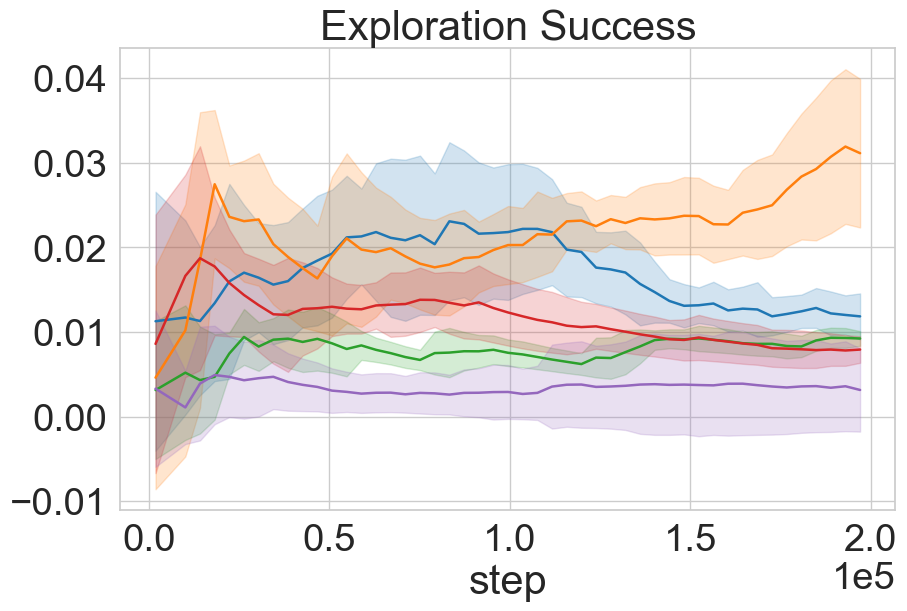}
  \includegraphics[width=0.22\columnwidth]{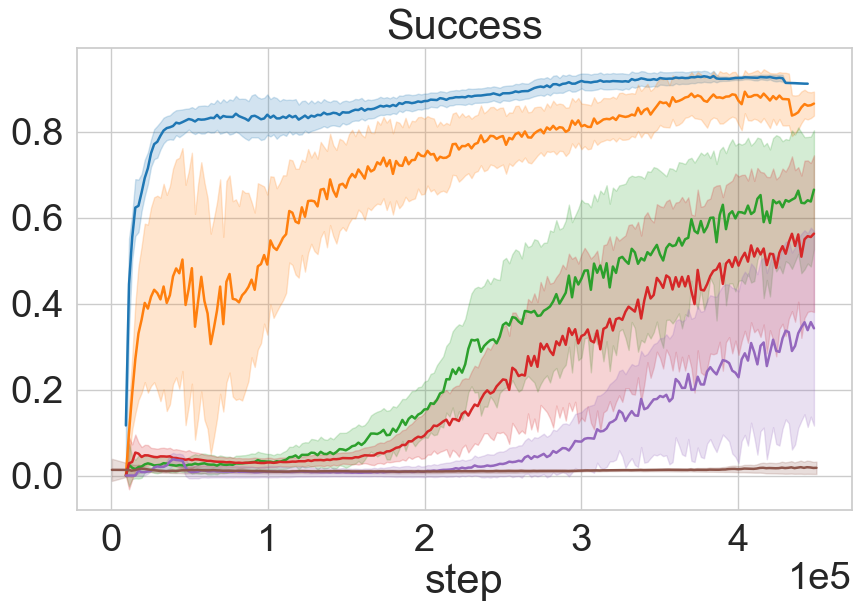}
  \includegraphics[width=0.22\columnwidth]{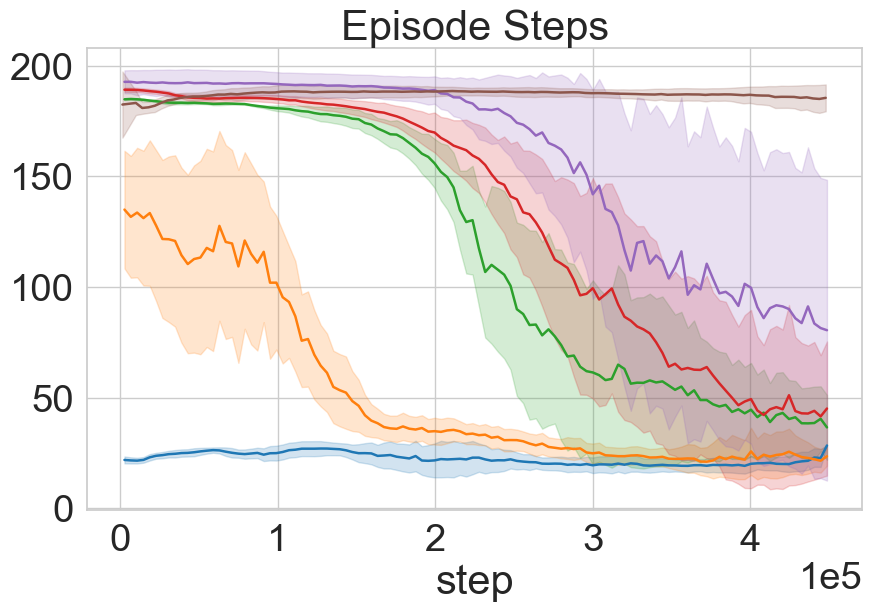}
  \includegraphics[width=0.22\columnwidth]{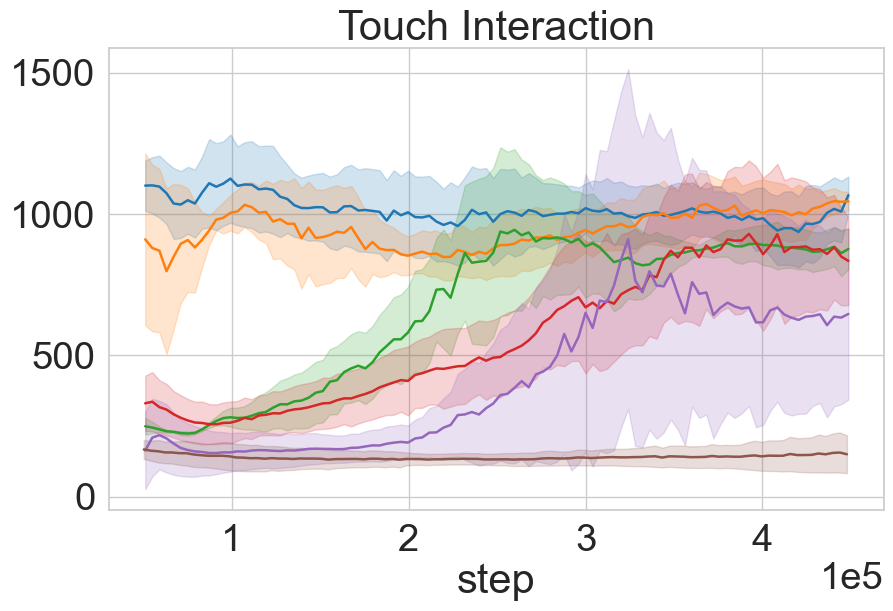}\\
  \includegraphics[width=0.7\columnwidth]{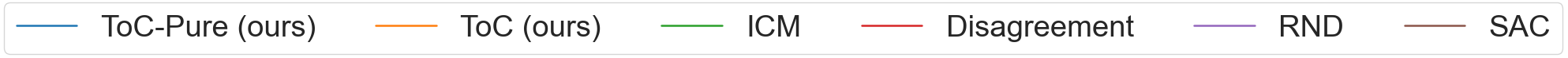}
  \caption{\textbf{MiniTouch Evaluations}. Each row in the figure outlines the performance of the \method{} variants and the baselines on a MiniTouch task. Each column marks performance on the four specified evaluation metrics over a number of training steps on the x-axis expressed in $1e^5$. The results are averaged across 5 random seeds and shaded areas represent mean ± one standard deviation while darker line represents mean. In the majority of the tasks, \method{} agents attain success in the \emph{exploratory} phase with no external reward (see text). \emph{Note}: We exclude the single object playing task as success in this task is equivalent to object interaction as depicted in Figure.~\ref{fig:single_object_variance}.}
  \label{fig:all_tasks}
\end{figure}

\section{Results and Discussion}
\method{} and baselines were trained on a Panda robot agent~\cite{coumans2019} for one million steps. In the \emph{exploratory} phase of the training, we pre-train our method only with the curiosity-based intrinsic reward. We then progress to the \emph{adaptation}  phase. Also, note that across all tasks, \method{}-Pure is based purely on cross-modal prediction, while \method{} includes visual forward prediction reward in addition.




Figure ~\ref{fig:single_object_variance} shows results on the basic task of playing with a single object. Since single object interaction does not have explicit goal states to evaluate, we instead measure the agent's ability to constantly engage and play with the underlying object. \method{}-Pure displays four times better interaction with the underlying object when compared to SAC (see Figure \ref{fig1:sub2}). Also, the plot depicting object movement metric in  Figure \ref{fig1:sub1} indicates that \method{} is slightly more dynamic in the interaction in contrast to \method{}-Pure. Note from the plots that there is a trade-off when using \method{} and \method{}-Pure between constant interaction (i.e. touch interaction performance) and object movement dynamics. 
Collecting a variety of such interesting data during the \emph{exploratory} phase helps the agent in terms of sample efficiency while solving the downstream tasks.  Figure ~\ref{fig:single_object_variance_door} in the supplemental material section shows similar plots for door opening task. We also include short video clips displaying qualitative performance on each of the tasks in the supplementary material.
\begin{figure}[t]
  \centering
  \includegraphics[width=0.22\columnwidth]{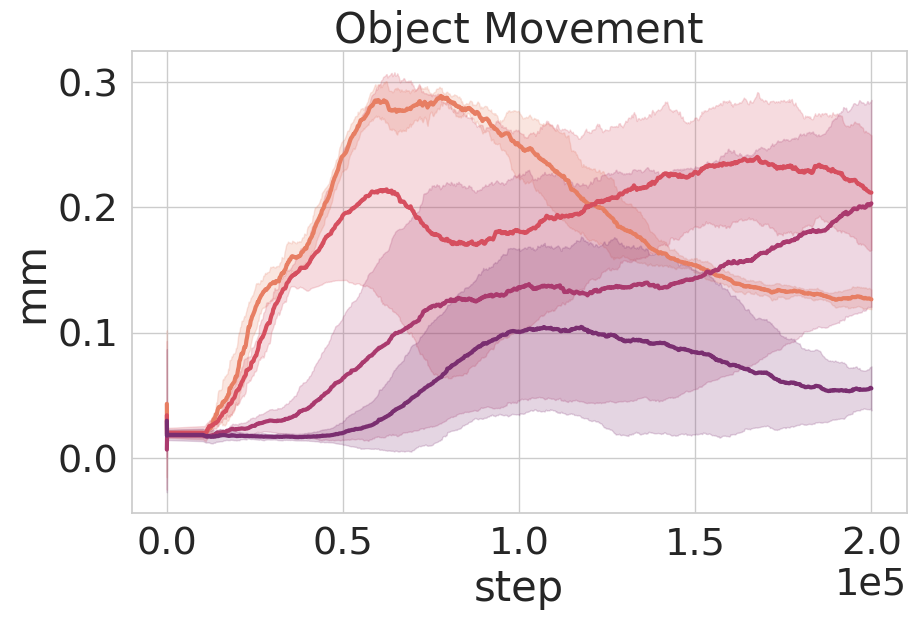}
  \includegraphics[width=0.21\columnwidth]{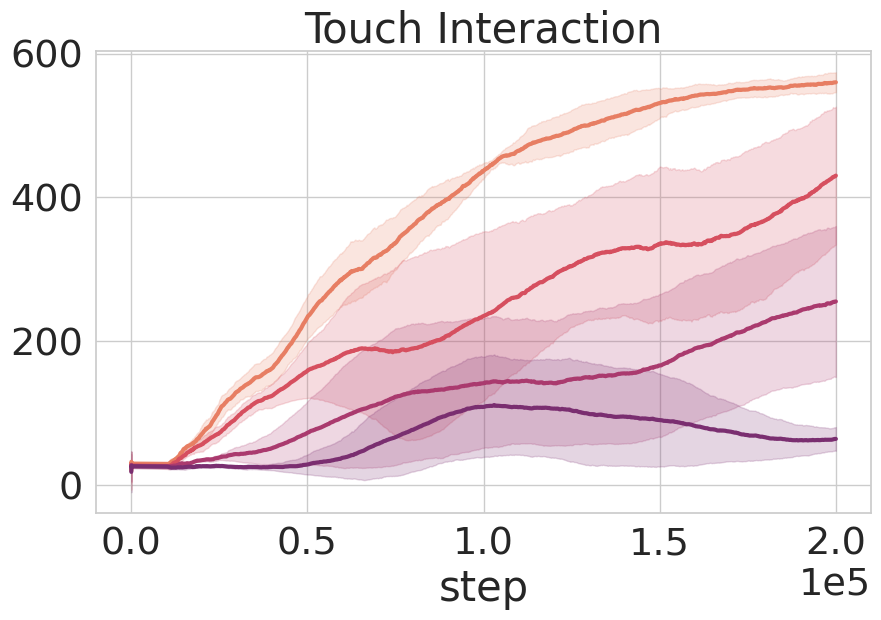}
  \includegraphics[width=0.225\columnwidth]{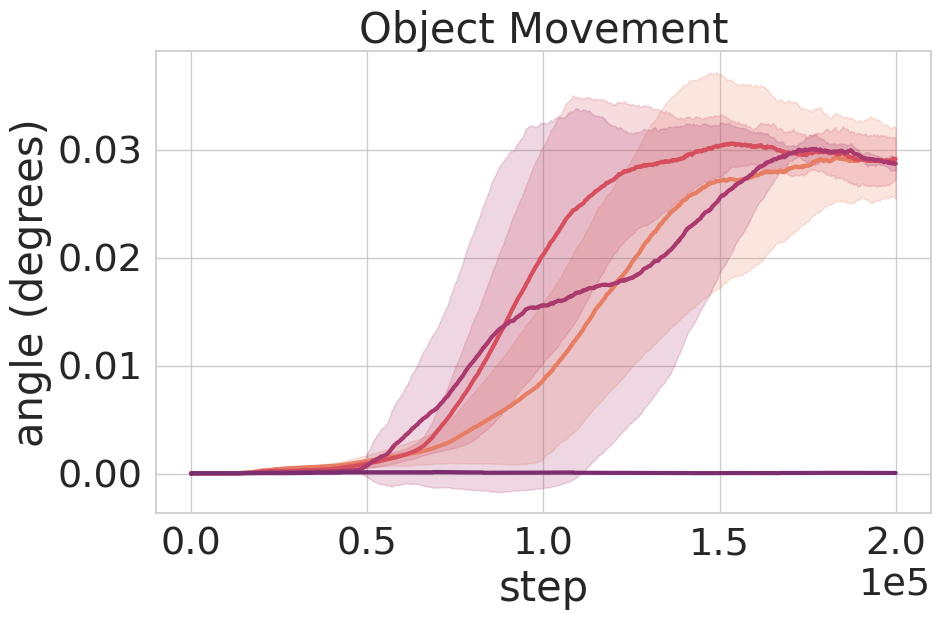}
  \includegraphics[width=0.21\columnwidth]{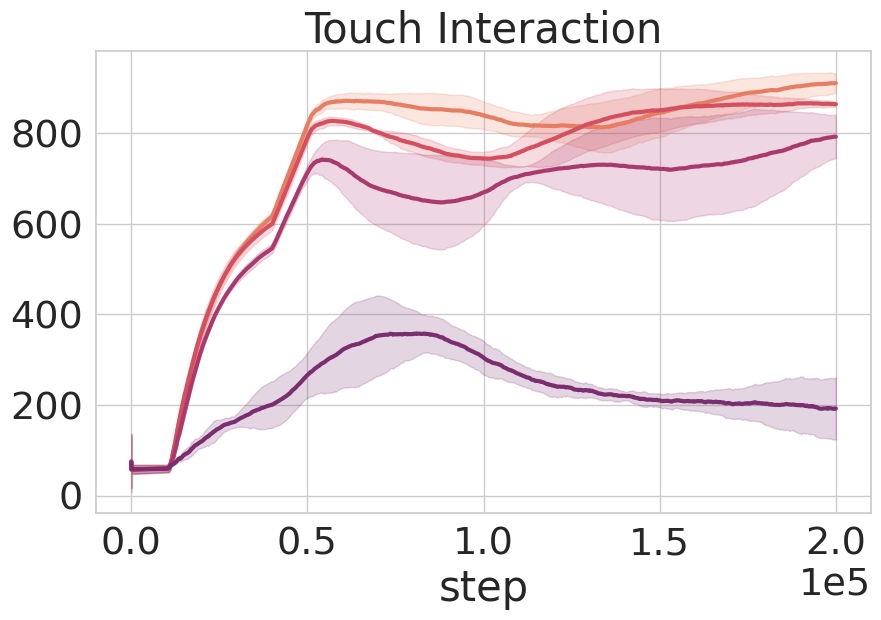}\\
  \includegraphics[width=0.35\columnwidth]{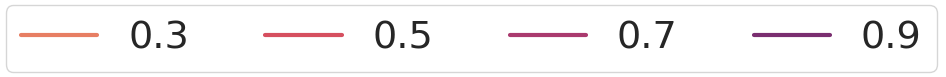}
  \caption{\textbf{Forward Prediction weight.}  Performance on Touch interaction and  Object movement metrics evaluated on the playing task and Open-door task for different forward prediction weightings sampled $\sim [1, 0]$. Large weight (darker plot) favors object-movement towards the end of training, where as smaller weight improves touch-interaction. Middle value in the sweep range (\emph{e.g.} 0.5) balances the trade-off.}
  \label{fig:ablation_forwardweight}
\end{figure}

We compare \method{} and \method{}-Pure with SAC and state-of-the-art vision-based curiosity baselines on the remaining downstream tasks in MiniTouch. From Figure ~\ref{fig:all_tasks} it is evident that \method{} and \method{}-Pure perform better than SAC in all the tasks and better than the vision-based curiosity models in the majority of the tasks. Using SAC alone hinders the performance and is often unable to solve any of the three tasks. This is not surprising since the model is not motivated enough to collect  diverse and useful data through interaction. ICM performs better than \method{}-Pure in the pushing task, however, \method{} dominates in performance by about 15\%. Recall that the goal is not just to succeed but to help attain success in a sample efficient manner in fewer steps. The results support our hypothesis that cross-modal curiosity enables an RL agent to succeeds at an early stage in training and often without any external reward. Similarly, our model outperforms on the opening task without external reward. However, \method{} initially has lower success compared  to RND, but surpasses towards the end. Although \method{} and \method{}-Pure attain similar success in the pick-up task towards the end of the training, it is compelling to note that \method{}-Pure attains faster convergence. This is because the picking task requires constant touch interaction (where \method{}-Pure has an advantage), as opposed to diverse object movement. In the following section, we
 examine the role of the forward objective term on touch interaction and possible ways to encode agent’s observation in our ablation studies.

\subsection{Ablations}\label{ablation}

\begin{table*}[t]
\vspace{-0.1cm}
    \centering
    \renewcommand{\arraystretch}{1.2}
    \setlength{\tabcolsep}{2pt}
    \begin{adjustbox}{width=0.94\columnwidth,center}
    \begin{tabular}{cccc|ccc|ccc|ccc}
        \toprule
     \makecell{Metric } & \multicolumn{3}{c}{\makecell{Pushing}} & \multicolumn{3}{c}{\makecell{Open Door}} &\multicolumn{3}{c}{\makecell{Pick-up}}&\multicolumn{3}{c}{\makecell{Playing}} \\
        \cmidrule(r){1-13}
      & \method{} & \method{}-\emph{fut} & ICM & \method{} & \method{}-\emph{fut}& ICM& \method{} & \method{}-\emph{fut}& ICM & \method{} & \method{}-\emph{fut}& ICM \\ \midrule
       \textbf{Exploration $\uparrow$}       & \textbf{0.403} & 0.291& 0.187 & \textbf{0.669} & 0.355  &0.083& \textbf{0.063} &0.051&0.013 &- &-&- \\
             \textbf{Success $\uparrow$}           & \textbf{0.733} & 0.678 & 0.597 & \textbf{0.983} & 0.571  &0.114& \textbf{0.891} &0.825 &0.780& -&-&-\\
    \textbf{Episode steps $\downarrow$}.    & \textbf{57.84} & 87.61 & 95.24 & \textbf{23.34} & 97.10  &199.3& \textbf{30.54} &33.77 &42.19& -&- &-\\
    \textbf{Touch-interaction $\uparrow$} & \textbf{247.79} & 210.11 & 202.66 & \textbf{600.1} & 287.97 &43.56& 980.7 &\textbf{984.2} &952.3 &\textbf{388.15}& 267.021&  63.31  \\
        \bottomrule
    \end{tabular}
    \end{adjustbox}
     \caption{\textbf{Touch-based future prediction} This table compares the mean evaluations for \method{} and \method{}-Pure on all the four tasks emphasizing the importance of the cross-modal association(see text). We omit measuring success and episode steps for playing task since success there is equivalent to object-interaction and has no explicit goal.}.
     \label{tab:results_ict_touch}
\end{table*}

\textbf{Importance of forward objective.}\label{forward_ablation}
Visual forward prediction $L_{fdm}$ plays a handy role when it is used in the right proportion. Intrinsic reward is a weighted combination of cross-modal prediction and forward prediction as defined in Eq.\ref{eq:final-reward}.  Figure \ref{fig:ablation_forwardweight} illustrates the performance of the model with different levels of emphasis on the forward loss term, with $\lambda$ uniformly sampled between 0 and 1. Higher weights indicate that the future prediction dominates force/torque prediction. We notice that both larger and smaller values of $\lambda$ hurt the overall performance. Larger weight leads to more object movement but hurts the robot's constant touch-interaction with the underlying objects, while smaller value leads to inactive behavior while satisfying the objective as shown in Figure.\ref{fig:ablation_forwardweight}.

\begin{figure}
 \begin{subfigure}{0.54\linewidth}
\centering
\includegraphics[width=0.41\columnwidth]{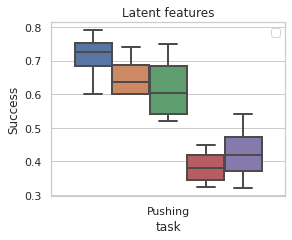}
\includegraphics[width=0.412\columnwidth]{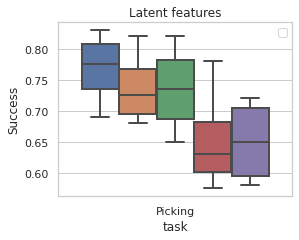}
\includegraphics[width=0.155\columnwidth]{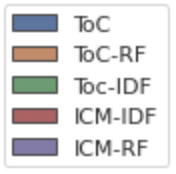}
\caption{Latent  feature space ablations}
\label{fig:latent_sub}
\end{subfigure}%
\begin{subfigure}{0.38\linewidth}
\centering
\includegraphics[width=0.46\columnwidth]{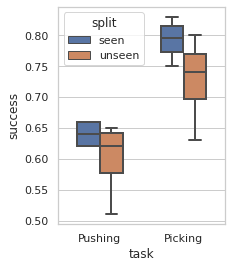}
\includegraphics[width=0.46\columnwidth]{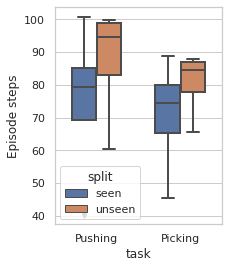}
\caption{Generalization to novel objects.}
\label{fig:generalization_unseen}
\end{subfigure}%

    \caption{(a) Comparing different latent space performance  on \emph{pushing} and \emph{grasping} tasks. Features learned via touch prediction perform better than those learned using IDF or Random. (b)Generalization performance on \emph{success} and \emph{episode steps} for the \emph{pushing} and \emph{picking} task. }
    \label{fig:latent}
\end{figure}%

\textbf{Touch-based future prediction.} The goal of this experiment is to strengthen the argument of cross-modal association. While conducting experiments, it is important to deduce information of one modality from another modality in a related manner than simply adding another modality on top of visual information. We created  an additional touch-based baseline, \method{}-\emph{future}, where in addition to the visual future prediction model
we include the touch-based future prediction model. The touch-based future prediction model takes a touch vector as input and predicts the touch vector for the next time step. Table~\ref{tab:results_ict_touch} compares  \method{} and \method{}-\emph{future} on MiniTouch tasks and we observe that \method{}-\emph{future} is  better than ICM in general but compares below \method{}.

\textbf{Robustness to diverse shapes.} 
It is desirable for an agent to be able to handle diverse shapes  in order to be robust across arbitrary manipulation settings.  We study this using an environment in which an object is sampled from a thousand procedurally generated objects. The objects are dissimilar with respect to shape and mass but are sampled from the same generative distribution. Out of 1000 different objects, 800 of them are used in the training phase, and we evaluate the agent's effectiveness on the remaining 200 unseen object shapes. Figure ~\ref{fig:generalization_unseen} shows  touch interaction and object movement evaluations for a single object exploration task. The results validate that our model generalizes to unseen object configurations.

\textbf{Latent features for forward dynamics.} Choosing  ideal embedding space for decoding the touch vector (Figure \ref{fig:1a}) and for predicting future state (Figure. \ref{fig:1b}) is important. Existing approaches rely on a pretext inverse dynamics task of predicting the agent’s action given
its current and next states~\cite{Pathak}. Another simple yet strong method is to use features from a random but fixed initialization of the encoder~\cite{Burda}.
In our work, we learn the features by leveraging the self-supervised pretext task of predicting one modality from the other. Figure.\ref{fig:latent_sub} compares (1) encoder, i.e. learned through cross-modal prediction(\method{}) (2) random feature encoder(\method{}-RF) (3) encoder learned through IDF task (\method{}-IDF). In each case, the decoder network is optimized through touch prediction. We observe that the random features variant is stable and effective on both \method{} and ICM models.




\section{Conclusion}

 Building self-seeking RL agents that can leverage cross-modal associations could be a key research direction in developmental robotics. Sense of touch is ubiquitous in human routine. Borrowing insights from how infants learn to explore out of curiosity by touching what they visually
perceive,  we formulated  Touch-based Curiosity (\method{}) that is aimed at encouraging exploration via haptic interaction with the environment. We demonstrated in this work, that by involving additional modalities, the performance of curiosity-based systems on downstream tasks can be increased. 
We observed increased interaction with target objects, and presented evidence that \method{} learns to solve the MiniTouch benchmark tasks in an efficient manner while vanilla RL algorithms and vision-based curiosity formulations struggled.
Force/torque sensing is widespread in the lab and industrial robots but while there are plenty of robotic benchmarks, we believe that tactile feedback is an under-explored modality and by releasing our benchmark, we hope to enable future research in this exciting area.



\section*{Acknowledgements}
 We would like to thank Deepak Pathak, Glen Berseth, Edward Smith and  Krishna Murthy for valuable feedback and insightful comments. 


\bibliography{example}  

\newpage
\section{Supplementary Material}\label{sec:appendix}
  
In this Supplementary Material section, we provide additional details concerning various elements  which could not be elaborated on in the main paper. We begin with a detailed
description of the proposed MiniTouch benchmark. We outline the action space, state space, and reward structure for each of the tasks. This is followed by a closer look into the various aspects of the touch-control module
including architectures, experimental procedures, hyper-parameters, and additional results. We
include an ablation experiment at the end that investigates the usefulness larger exploration phase while solving each of the tasks.

\begin{figure}
    \centering
    \includegraphics[width=0.27\textwidth]{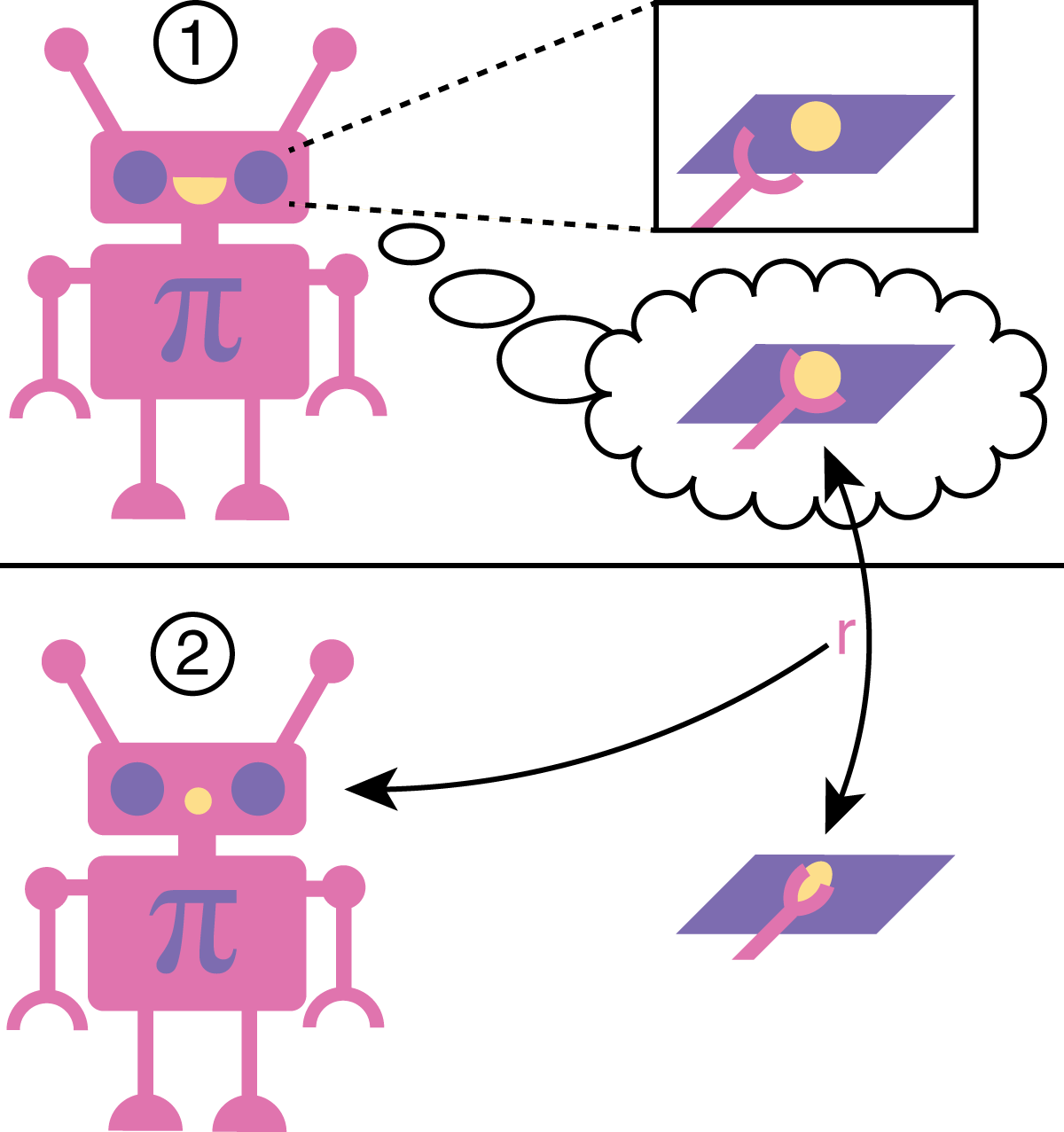}
    \caption{\textbf{Touch-based Curiosity (\method{}) Overview} \emph{Top}: An agent perceives a scene visually and anticipates the force/torque (FT) sensation of interacting with an object. \emph{Bottom}: The object interaction leads to an unexpected FT sensation, which gives a positive reward to the policy, leading to an exploration policy that is seeking interactions that are haptically surprising. The agent's experiences gained in this way are later relabeled to become task-specific.}
    \label{fig:overview}
\end{figure}

\subsection{MiniTouch Tasks}\label{sec:benchmark_sec}
 
 This section describes the cross-modal benchmark of simulated manipulation tasks, which we make use of and release as part of the paper. Unlike these prior simulation benchmarks, we particularly focus on providing a platform where one can use cross-modal information to solve diverse manipulation tasks.  Existing benchmarks~\citet{metaworld} do not include touch modality. An overview of the tasks in MiniTouch is outlined in Figure \ref{fig:test-tasks}. These tasks are  built off Pybullet~\cite{coumans2019} physics engine and contain different scene setups for each of the four tasks. The details of each of these tasks are further expanded. All the four tasks are compiled together as a ``MiniTouch" benchmark suitable for evaluating interaction-based algorithms.

\textbf{Playing:} This environment is intended as a toy task to evaluate interaction frequency and does not feature any reward beyond interaction count. A cube is placed in a random position on a table at each episode. The agent needs to localize and interact with the cube.

\textbf{Pushing:} In this task, the agent needs to push an object placed randomly on a table to a target (visually indicated as a gray cube). The object position is sampled uniformly in polar coordinates around the target object (i.e. angle 0 to 360 degrees, distance 10 to 20 centimeters). The end effector's start position is sampled in the same way as the target position. In addition, the orientation of the gripper is fixed to be perpendicular to the ground all the time.
The robot agent succeeds and receives a reward of +25 if the distance between the cube and the target object is less than 7 centimeters. A new episode starts if the agent succeeds. The environment also restarts if the cube is placed or pushed beyond a predefined bounding box comprising of acceptable positions on the table (\textit{i.e.} positions that can be reached by the robot hand).

\textbf{Opening:} A cabinet with a door is randomly placed in reach of the agent. The goal is to find the door handle and open the door. The gripper orientation is fixed to point its fingers towards the door, parallel to the ground. For this task, the fingers are discretized to be open or closed. In addition, a fifth element is added to the action vector to control the yaw (relative rotation of the end-effector) to be able to approach the door. The robot succeeds and receives a reward of +25 when the angle of the door opening reaches thirty degrees or higher. Similar to the pushing task, a new episode starts if the agent reaches the goal.

\textbf{Pick-up:} In this environment, the agent needs to grasp and lift a randomly placed object. The agent's goal is to lift the object  5cm above the table. The agent receives a reward of +25 upon success. The object is placed uniformly randomly on a table. Similar to the Opening task, the end effector opening/closing is discretized, meaning when its internal continuous variable is below a threshold, the gripper closes, otherwise it remains open.

All of the tasks are implemented in the Pybullet physics engine~\cite{coumans2019}, which is a free and open-source library that enables fast simulation.
\subsection{Code}

\subsubsection{\emph{MiniTouch} Library:}
The task environment used in the experiments is packaged and released as a python library\footnote[1]{https://github.com/ElementAI/MiniTouch} that can be easily plugged into the training code. Setup instructions, code, and other details can be found in the README file included in the repository.

\subsubsection{ToC and baselines:}
All of our models are written in the open-source PyTorch~\cite{pytorch} library. We open-source the implementation of our approach\footnote[2]{https://github.com/ElementAI/ToC}. The code trains a ToC exploration agent in the environment. A README file in the repository contains the setup instructions and acts as a starting point to reproduce the experiments on each of the \emph{MiniTouch} tasks.

We used the following open-source implementations for the baselines. We were able to reproduce the results from their papers before attempting to use them as baselines for our model:

\textbf{ICM}\\ 
https://github.com/openai/large-scale-curiosity 

\textbf{Disagreement}:\\
https://github.com/danijar/dreamerv2\\
https://github.com/pathak22/exploration-by-disagreement

\textbf{RND}:\\
https://github.com/openai/random-network-distillation

\subsection{Experimental Details}\label{suppl_experimental}

\textbf{State space:} The input states are a combination of visual and touch vector input. The visual input is a grayscale rendering of the scene with dimension $84\times84$, pre-processed similarly to~\citet{Mnih}. Image observations are captured from a static camera overlooking each scene. 
The touch vector input is composed of the 3-dimensional end-effector position, 2-dimensional finger position, ranging from 0 to 1, each denoting how far apart each finger is, and the 6-dimensional force/torque values. 
In total, the touch vector is 11-dimensional, $S\in\R^{11}$.\\
\textbf{Action space:} Actions are expressed as 4-dimensional continuous vectors. 
The first three elements describe the desired offset for the end effector at the next timestep based on its current position. The last dimension controls the relative desired distance between the two fingers at the next timestep. 

\paragraph{Training}
We use SAC~\cite{haarnoja2018soft} as the optimizer for our agent. Our training is composed of two phases as described in the main paper. (i)Exploration phase, (ii)Downstream task phase. In the exploratory phase (curiosity part) agent is trained using our intrinsic reward alone.  In the task phase, network weights are seeded from the ones in the exploratory phase. We also retain the replay buffer from the exploratory phase in the downstream task phase.  Duration of the exploration phase can be adjusted in the code using a hyper-parameter \emph{stop-curiosity}. We have two hyper-parameters that change between the two phases, (i) $\alpha$ and (ii) the learning rate of the SAC algorithm. Details of hyper-parameters used for our experiments are outlined in Table \ref{table:SAC_hp} and Table \ref{table:visual_network}.

\begin{table}[ht]
\centering
\begin{tabular}{l|l}

\hline
\multicolumn{2}{c}{\textbf{SAC pretraining and training hyperparameters}} \\

\hline
  \makecell{Parameter type}   & Value \\ 
\hlx{vhv}

optimizer& Adam \\
Visual network & Table \ref{table:visual_network} \\
 learning rate  & $3 . 10^{-5}$ \\
 number of samples per minibatch & 128 \\
 reward scale & $100$ \\
 replay buffer size & $10^{6}$ \\
 number of hidden units per layer & 128 \\
 number of hidden layers (all networks) & 2 \\
 activation & LeakyReLU \\
 discount factor & $0.99$\\
\hline
\end{tabular}
\caption{SAC parameters during pretraining and training.}
\label{table:SAC_hp}
\end{table}













\begin{table}[H]
\centering
\begin{tabular}{l|l|l|l|l}

\hline
\multicolumn{4}{c}{\textbf{Encoder network}} \\

\hline
  \makecell{Layer}   & \makecell{Number of outputs \\} &  Kernel size &Stride&Activation function\\
\hlx{vhv}

 Input $x$  & $84 * 84 * 1$ & && \\

Convolution & 20*20*32 & $8 * 8$&4&LeakyReLU\\
Convolution & 8*8*64 & $4 * 4$&2&LeakyReLU\\
Convolution& 4*4*124 & $3 * 3$&1&LeakyReLU\\
Convolution& 2*2*256 & $2 * 2$&1&LeakyReLU\\
Fully-connected& 256 & 1&  &LeakyReLU\\
\hline
\end{tabular}
\caption{Visual network.}
\label{table:visual_network}
\end{table}

\subsection{Object Interaction}
\begin{figure*}
\centering
\begin{subfigure}{0.3\linewidth}
\includegraphics[width=0.99\columnwidth]{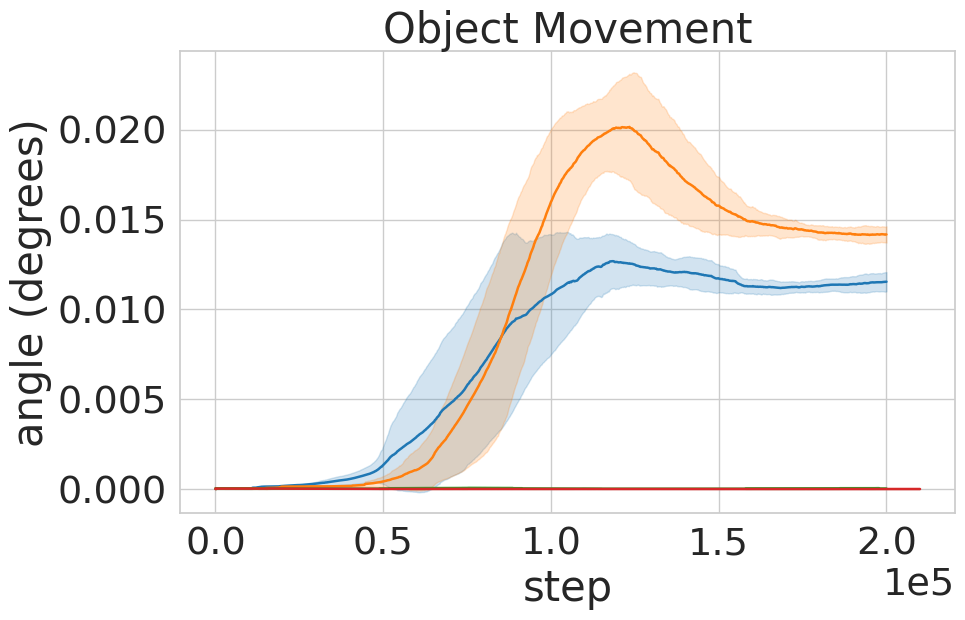}
\caption{}
\label{fig1:sub1_door}
\end{subfigure}%
\begin{subfigure}{0.3\linewidth}
\includegraphics[width=0.99\columnwidth]{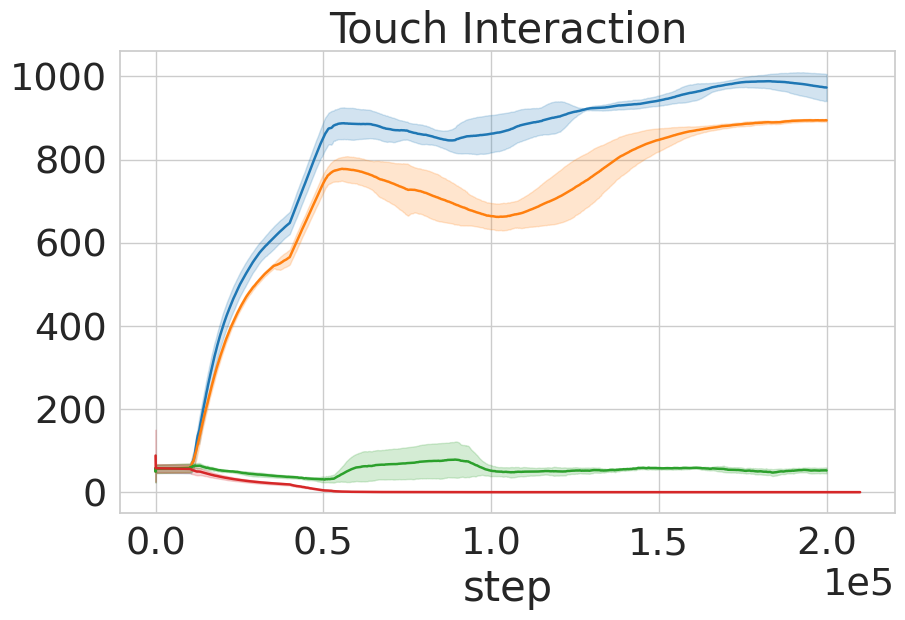}
\caption{}
\label{fig1:sub2_door}
\end{subfigure}\\
\begin{subfigure}{0.5\linewidth}
\centering
  \includegraphics[width=0.9\columnwidth]{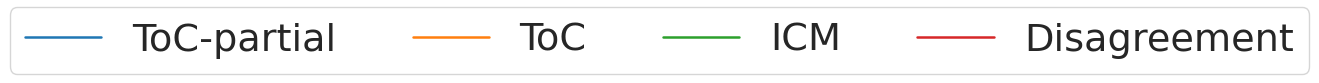}
  \end{subfigure}%
\caption{\textbf{Object Interaction for Opening task.}  (a) shows the average variance in door angle across the entire episode (note that the absolute variance is low but corresponds to successful door openings towards the end of training for \method) and in (b), we count the number of touch interactions in the same task.}
\label{fig:single_object_variance_door}
\end{figure*}

As touched up in the experiments section of the main paper, Figure \ref{fig:single_object_variance_door} depicts the touch interaction and door movement metrics for the Opening task. We make a similar observation to the result showed in Figure \ref{fig:single_object_variance} for the Playing task. A higher touch-interaction need not indicate better object-movement. Agent can resort to constantly engaging with
the object in a passive manner. \method{} collects rich  interaction data during the exploratory phase and helps the agent in terms of sample efficiency while solving the downstream tasks.

\subsection{Ablation}
\begin{figure}
  \centering
  \includegraphics[width=0.25\columnwidth]{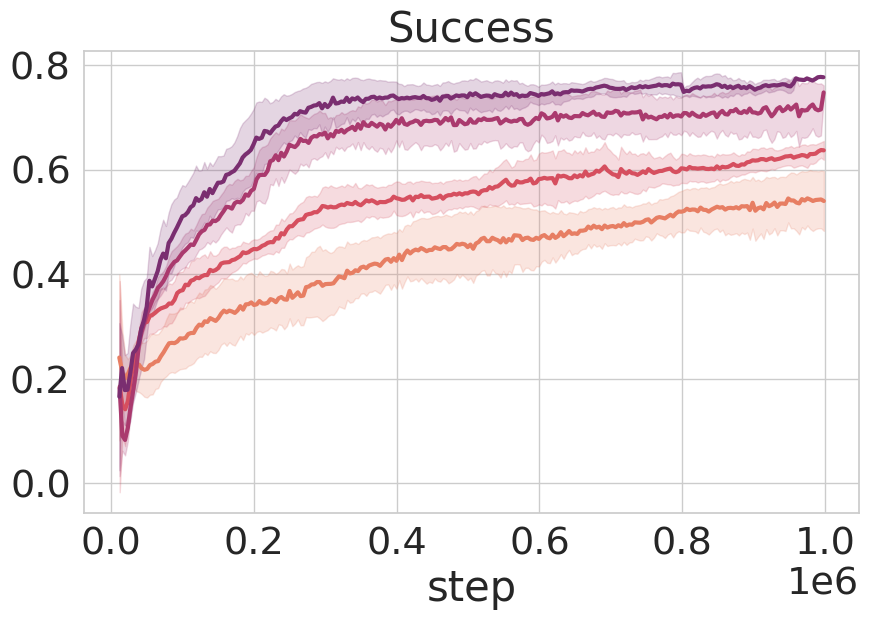}
  \includegraphics[width=0.25\columnwidth]{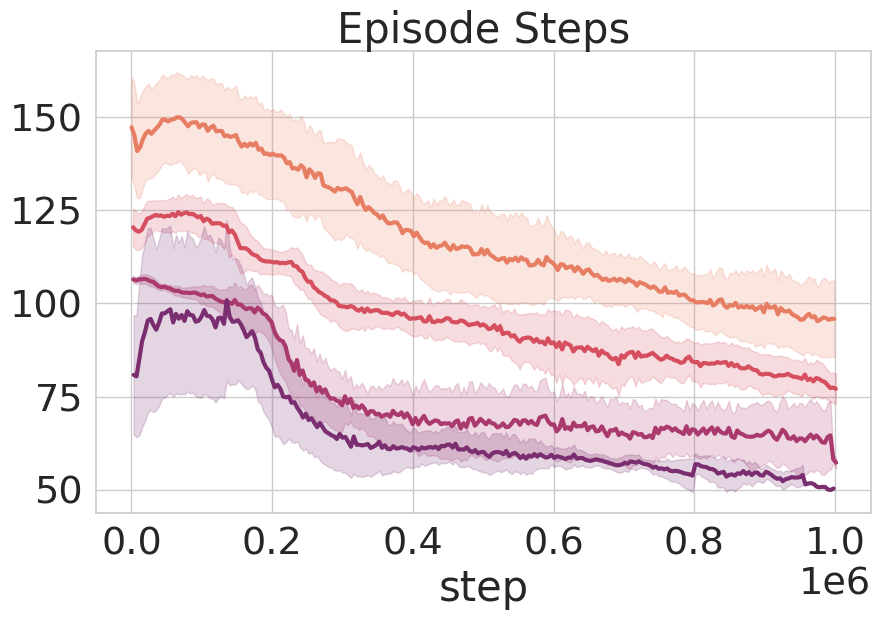}\\
  \includegraphics[width=0.25\columnwidth]{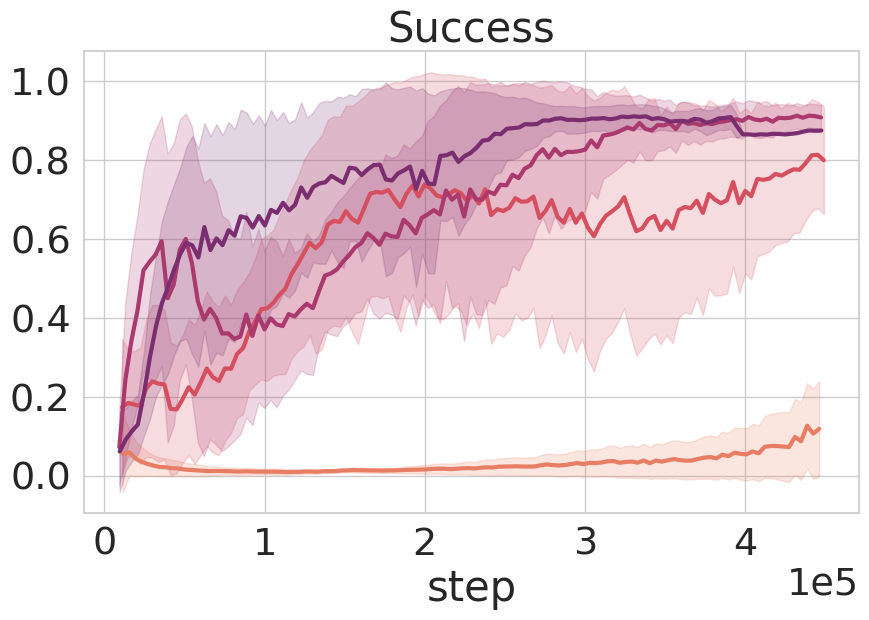}
  \includegraphics[width=0.25\columnwidth]{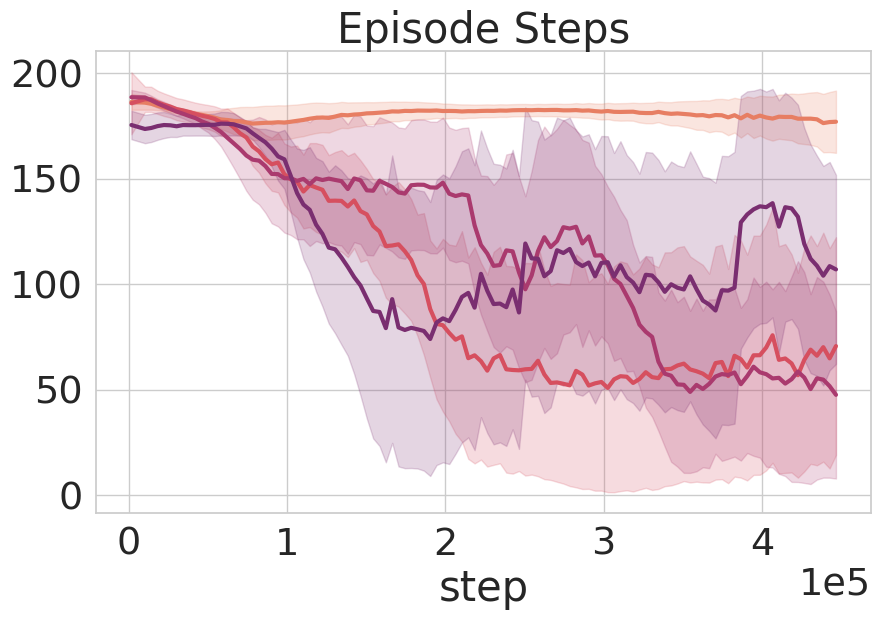}\\
  \includegraphics[width=0.25\columnwidth]{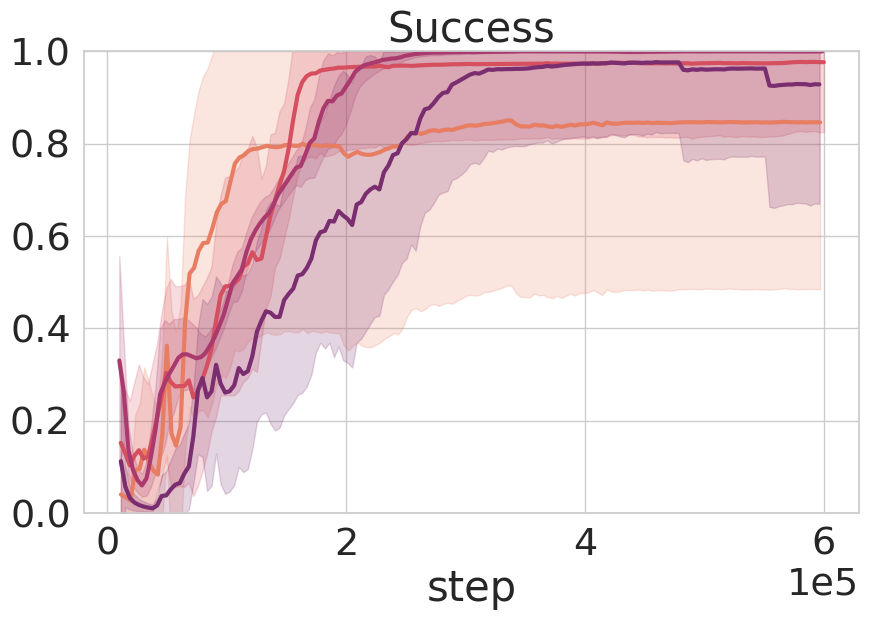}
  \includegraphics[width=0.25\columnwidth]{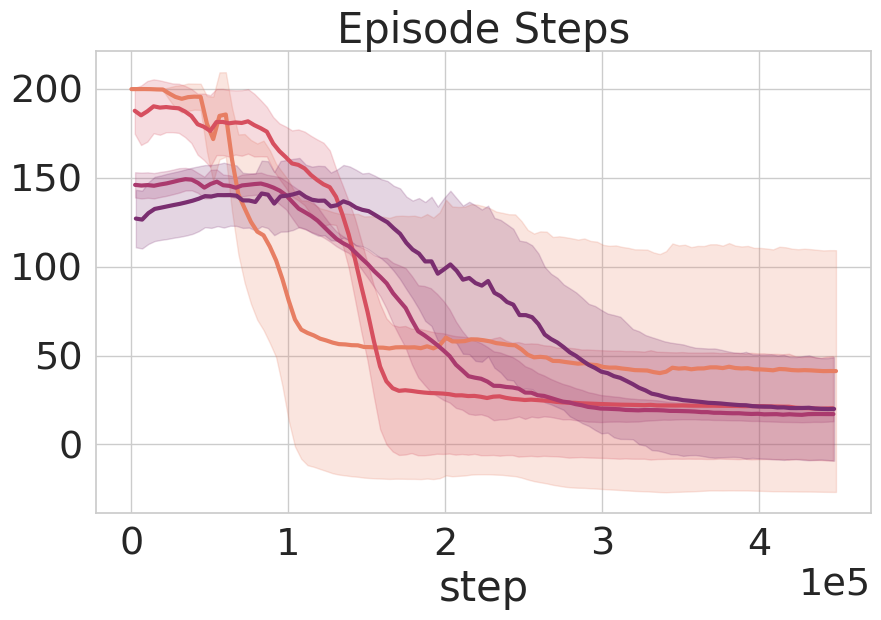}\\
  \includegraphics[width=0.55\columnwidth]{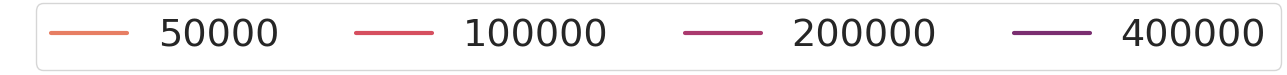}
  \caption{\textbf{Longer exploratory phase helps}  Success and episode steps evaluations on different tasks for different lengths of \emph{exploratory} phase. Darker shading indicates longer exploration.}
  \label{fig:ablation_stopcuriosity}
\end{figure}

\textbf{Does longer exploration help?}
We observe that having a longer exploratory phase of training with intrinsic reward alone usually benefits the overall performance. We can observe from Figure ~\ref{fig:all_tasks} that \method{} attains decent success in the exploratory phase without any external reward on most of the tasks. This is because it encourages better associations and a larger collection of interesting configurations in the replay.
The effect of the exploratory step is further studied and the results on all of the downstream tasks with different duration of exploration are compiled in Figure ~\ref{fig:ablation_stopcuriosity}. We depict the episode convergence steps and success rate to visualize the trend as expected.

\end{document}